\documentclass[times,authoryear]{elsarticle}

\usepackage{framed,multirow}
\usepackage{comment}

\usepackage{amssymb}
\usepackage{latexsym}

\usepackage[switch]{lineno}

\usepackage{url}
\usepackage{xcolor}
\usepackage{listings}
\usepackage{amsmath}

\usepackage{algorithm}
\usepackage{subcaption}
\usepackage{algpseudocode}
\usepackage{graphicx}
\usepackage{multirow}
\usepackage{booktabs}
\definecolor{newcolor}{rgb}{.8,.349,.1}

\usepackage[citebordercolor=white]{hyperref}

\providecommand{\snm}[1]{#1}

\begin{document}


\begin{frontmatter}

\title{Current state of the multi-agent multi-view experimental and digital twin rendezvous (MMEDR-Autonomous) framework}

\author[1]{Logan \snm{Banker}\corref{cor1}}
\author[2]{Michael \snm{Wozniak}}
\author[1]{Mohanad \snm{Alameer}}
\author[1]{Smriti Nandan \snm{Paul}}
\author[1]{David \snm{Meisinger}}
\author[1]{Grant \snm{Baer}}
\author[1]{Trevor \snm{Hunting}}
\author[1]{Ryan \snm{Dunham}}
\author[1]{Jay \snm{Kamdar}}

\affiliation[1]{organization={Department of Mechanical and Aerospace Engineering},
                addressline={Missouri University of Science and Technology},
                city={Rolla},
                postcode={65409},
                state = {MO},
                country={United States}}

\affiliation[2]{organization={Purdue University},
                addressline={610 Purdue Mall},
                city={West Lafayette},
                postcode={47907},
                country={USA}}


\begin{abstract}
As near-Earth resident space objects proliferate, there is an increasing demand for reliable technologies in applications of on-orbit servicing, debris removal, and orbit modification. Rendezvous and docking are critical mission phases for such applications and can benefit from greater autonomy to reduce operational complexity and human workload. Machine learning-based methods can be integrated within the guidance, navigation, and control (GNC) architecture to design a robust rendezvous and docking framework. In this work, the Multi-Agent Multi-View Experimental and Digital Twin Rendezvous (MMEDR-Autonomous) is introduced as a unified  framework comprising a learning-based optical navigation network, a reinforcement learning-based guidance approach under ongoing development, and a hardware-in-the-loop testbed. Navigation employs a lightweight monocular pose estimation network with multi-scale feature fusion, trained on realistic image augmentations to mitigate domain shift. The guidance component is examined with emphasis on learning stability, reward design, and systematic hyperparameter tuning under mission-relevant constraints. Prior Control Barrier Function results for Clohessy-Wiltshire dynamics are reviewed as a basis for enforcing safety and operational constraints and for guiding future nonlinear controller design within the MMEDR-Autonomous framework. The MMEDR-Autonomous framework is currently progressing toward integrated experimental validation in multi-agent rendezvous scenarios. 
\end{abstract}


\end{frontmatter}

\section{Introduction}

Spacecraft rendezvous and docking are critical capabilities that enable a broad range of spaceflight missions, including active debris removal (ADR), in-space servicing, assembly, and manufacturing (ISAM), and orbit modification. While previous rendezvous and docking missions relied heavily on human supervision and intervention \citep{ZimpferSurvey}, such approaches are ultimately unscalable due to safety risks, operational complexity, and high costs. This limitation is of growing concern as the number of resident space objects (RSOs) continues to increase and orbital traffic becomes increasingly difficult to manage. In conjunction with this, the Office of the Inspector General recognizes that preventive measures alone are no longer sufficient to manage the accelerating growth of orbital debris, and that active debris removal technologies must advance to guarantee long-term orbital sustainability \citep{NASAAudit}. Together, the operational, regulatory, and environmental pressures point towards future missions requiring docking technologies that are both autonomous and scalable. 

In response to these needs, the Multi-Agent Multi-View Experimental and Digital Twin Rendezvous (MMEDR-Autonomous) framework is being developed as an architecture for designing, testing, and validating autonomous guidance, navigation, and control (GNC) solutions. It specifically addresses the GNC requirements of future ADR and ISAM for multi-agent rendezvous missions. The framework emphasizes algorithmic robustness, system-level integration, and feasibility of deployment on compact spacecraft. CubeSats, in particular, impose strict limitations on size, power, propulsion capability, and onboard computational resources. Constraints of this nature require GNC subsystems that are computationally efficient and capable of maintaining safe operation under mission conditions such as relative-state uncertainty and velocity limits.

Traditionally, spacecraft rendezvous systems are organized around three core functions: guidance, navigation, and control. The guidance component generates feasible motion profiles, navigation provides state estimates from onboard sensing, and control executes the commanded motions while enforcing actuator and safety constraints. These systems have historically relied on established techniques such as Clohessy-Wiltshire (CW) relative dynamics, extended Kalman filtering, model-based guidance and control laws, and geometric pose estimation methods based on feature correspondences. Although such methods form a reliable foundation, their performance can degrade in the presence of multi-agent interactions, strongly nonlinear dynamics, and adverse sensing conditions. As a result, machine learning approaches, including reinforcement learning (RL), convolutional neural networks (CNNs), and multi-agent learning frameworks, have increasingly been explored as a means of improving autonomy and robustness in rendezvous and docking scenarios \citep{carletonRLAlgorithm, OnGroundVivek, SPNv2}. However, before these methods can be deployed as part of an integrated GNC architecture, their compatibility, computational demands, and stability characteristics must be carefully examined, particularly when operating on resource-constrained platforms such as CubeSats. Within this broader context, this work investigates learning-based methods within the GNC architecture of the MMEDR-Autonomous framework.

The first contribution of this work is the study of learning stability in reinforcement learning-based guidance algorithms. Much of the existing literature prioritizes reward shaping as the primary means of influencing learning behavior, often giving limited attention to the effects of hyperparameter selection, state definition, or the staged introduction of safety constraints. This work therefore examines reinforcement learning formulations with different state definitions and reward functions, the incremental introduction of safety constraints, and the configuration of a Bayesian optimization framework used for hyperparameter tuning.

The second contribution of this work is the development of a neural network architecture for satellite pose estimation using monocular imaging. The architecture combines a lightweight backbone \citep{MobileNetV3} with a direct regression pose task \citep{Deep-6DPose}, allowing 6D pose estimation to be learned alongside complementary tasks within a unified network. Data augmentation methods are incorporated into the training process to better support navigation functions in real testing scenarios, improving the accuracy of the perception module within the autonomous docking pipeline.

The third contribution of this work is the design and early development of a hardware-in-the-loop (HIL) facility intended for evaluating autonomous GNC algorithms within a controlled laboratory setting. Although the facility remains under active development, substantial progress has been made in establishing the requirements necessary for closed-loop validation. The testbed is being configured to accommodate cooperative multi-agent rendezvous scenarios with large RSOs under space-representative conditions. This capability supports the maturation of learning-based approaches prior to deployment.

The remainder of this paper is organized as follows: Section~\ref{sec:rw} reviews existing literature on guidance, control, and navigation methods for spacecraft rendezvous and docking, as well as prior experimental testbeds used for validation. Section~\ref{sec:ps} introduces the governing dynamics, defines the single-agent docking problem, and then extends the discussion to the multi-agent setting. Section~\ref{sec:as} presents the solution methods, including the reinforcement learning-based guidance approach, the optical navigation network, and the state-estimation method that accounts for delayed and asynchronous measurements. Section~\ref{sec:integration} describes the laboratory setup and the integration of framework subsystems. Section~\ref{sec:results} presents preliminary results from the MMEDR-Autonomous framework and their implications. Finally, Section~\ref{sec:conclusions} concludes the paper and outlines future research directions.

\section{Related Work} \label{sec:rw}

\subsection{Guidance}
The guidance subsystem is vital to rendezvous and docking missions because it supplies the desired signal (such as ideal position or velocity) to a controller that subsequently minimizes the error between the current state and ideal state through conventional control methods. Two notable methods of generating a guidance signal are numerical optimization and reinforcement learning.
\subsubsection{Numerical Optimization}
\indent One method of obtaining a guidance signal involves iterative/numerical optimization methods to minimize a performance index.  An example of this application is presented in \cite{Ventura2017}, which accomplishes minimum energy maneuvers by minimizing the following performance index \(J\) while satisfying terminal constraints:
\begin{align}
  J = \frac{1}{2}\int_0^{t_f}\left(\left(\frac{F_x^C}{m_C}\right)^2+\left(\frac{F_y^C}{m_C}\right)^2+\left(\frac{F_z^C}{m_C}\right)^2+\left(\frac{T_x^C}{L_{eq}m_C}\right)^2+\left(\frac{T_y^C}{L_{eq}m_C}\right)^2+\left(\frac{T_z^C}{L_{eq}m_C}\right)^2\right)dt
\end{align}
where \(L_{eq}\) is a length parameter to convert the control torque to an equivalent force, \(F_i^C\) and \(T_i^C\) are forces and torques for \(i=x,y,z\), and \(m_C\) is the mass of the chaser spacecraft. This minimum energy maneuver yields a continuous, quadratic, and differentiable performance index that can be solved using sequential quadratic programming (SQP) based algorithms. \newline
\indent Another optimization objective for minimizing fuel consumption while performing a rendezvous maneuver is formulated in \cite{wang2}. This work categorizes the phases of rendezvous and docking into a short-range search phase between 10 km and 500 m of relative separation, and a small-thrust final section of the journey from 500 meters relative distance to contact. The algorithm is designed to bring the chaser within 500 m of the target by  minimizing the following cost function:
\begin{align}\label{jcostfx}
    J = \phi(x(t_f)) + \frac{1}{2}\int_{t_0}^{t_f}(Q(x(t))+u(t)^TRu(t))dt
\end{align}
The terminal cost \(\phi(x(t_f))\) penalizes deviations from the desired final position and velocity, and is represented as \(\phi(x(t_f))=[\sqrt{x^2+y^2}-500]+\sqrt{\dot{x}^2+\dot{y}^2}\). Minimizing this term encourages \(||[x\ y]^T||=500\) and \(\vec{v}=0\). The terminal cost can be reformulated as reaching the target by removing the \(-500\) term, ultimately repurposing from a rendezvous cost function to a docking cost function. The integral term in the expression for \(J\) penalizes the control effort while encouraging motion that reduces the time required to reach the target. Note that \(R\) is a a diagonal matrix and \(u(t)\) is the control variable. Additionally, \(Q(x(t))=\frac{x\dot{x}+y\dot{y}}{\sqrt{x^2+y^2}}\) represents the velocity component in the direction of position. A large negative value implies quickly approaching the target, balanced out by the terminal constraint that incentivizes near-zero velocity. This cost function is formulated in a standard optimal control convention, demonstrating potential for standard optimization methods such as calculus of variations or dynamic programming. However, the optimization goal in \cite{wang2} was achieved via the Deep Deterministic Policy Gradient (DDPG) RL algorithm that employed (target) actor/critic networks to learn an ideal guidance signal from experience.

\subsubsection{Reinforcement Learning}
\indent Machine learning methods, such as Reinforcement Learning (RL), offer an alternative approach to generating guidance signals and can outperform traditional numerical methods, provided that stable learning is achieved. A chief illustrative example is the application of the distributed distributional deep deterministic policy gradient (D4PG) algorithm for learning a guidance strategy \citep{carletonRLAlgorithm}, which was experimentally validated via planar rendezvous \citep{carletonRLExperiment}. The methods in \cite{carletonRLAlgorithm} demonstrated that learning complex tasks through RL significantly reduces the engineering effort that would otherwise be necessary to handcraft solutions. While the force and torque commands for detumbling spinning space debris can be computed via the Udwadia-Kalaba equation as demonstrated in \cite{CarletonUdwadiaKalaba}, it can be challenging or impossible to formulate for more complex tasks. 

Multi-Agent Reinforcement Learning (MARL) is an extension of RL in which multiple agents learn simultaneously. Two possible categories of MARL are centralized and decentralized learning. In centralized learning, the agents actively share policy and/or gradient information through a coordinating entity which can expedite convergence at the trade-off of additional communication between agents. Conversely, decentralized learning involves each agent independently maintaining its own policy through local observations, mitigating the effects of communication failures but negatively impacting the learning potential of each agent due to a non-stationary environment. The use of multiple agents can increase performance and learning efficiency when multiple agents feed into the same replay buffer, as demonstrated in \citet{Barth-Maron2018-d4pg}. An example application of multiple learning agents is the use of decentralized reinforcement learning which converged to a scalable, cooperative algorithm in a multi-agent system of UAVs \citep{uav_marl_tracking}.  \newline
\indent In addition to learning a robust docking policy, additional behaviors may be learned via RL if driven by mission requirements. One example is learning collision avoidance, which is an imperative part of spacecraft trajectory planning. A Deep-RL controller is presented in \cite{collisionavoidanceRL}, training a 3-DoF manipulator to reach the grasping location of the target while avoiding collisions with other parts of the satellite. This paper demonstrated grasping of a single feature on a satellite while avoiding collision(s) with other parts of the satellite, using known satellite geometry to know whether there was a collision in training. The reward function formulation included a collision penalty term to deter the agent from trajectories resulting in collisions. Successful training was demonstrated over 3000 episodes using the DDPG algorithm to learn collision-free paths to the target. Another example involves the process of learning to detumble the target pre-capture, where the chaser agent(s) make intermittent contact with the tumbling target to reduce its angular velocity prior to the official capture as outlined in \cite{activedetumbling}. This process demands several collision-free trajectories to iteratively contact the target and reduce the target's angular velocity in the pre-capture phase. While the work in \cite{activedetumbling} did not leverage RL to learn pre-capture detumbling, this concept has potential for further study in RL applications.
\subsection{Control}

The controller should enable the agent to reach the desired state while preventing the agent from encountering undesirable states in between the initial and final states. The controller should provide an extra layer of safety to prevent collisions, sensor damage, or loss of target in the visual field. Two candidate architectures for the control system are velocity-based and acceleration-based control; namely, the control scheme may close the loop on velocity or acceleration, minimizing the error between the desired and actual state.

While PID controllers are popular in many fields, they are not used often in satellite rendezvous due to their sensitivity to noise and limited ability to handle complex nonlinear dynamic systems. Linear Quadratic Regulators (LQR) are a class of controllers utilized for rendezvous and docking missions due to the simplicity of application. LQR still requires a linear dynamical system but utilizes a quadratic cost function \citep{Mammarella}. Although it requires linearization for nonlinear systems, this controller provides increased robustness from the inclusion of the dynamic system into the control law. For noisy, nonlinear dynamical systems, Sliding Mode Control (SMC) can be utilized as an adequate method of control \citep{SMCtext}. Since SMC has a low response to disturbances, it is ideal in scenarios where the dynamic model does not accurately reflect the real dynamics. \cite{CapelloSMC} uses two SMCs, one first-order for position control and one second-order for attitude control. Model Predictive Control (MPC) is a popular form of optimal control used for solving controls problems in complex dynamic systems \citep{HartleyMPC}. MPC predicts future states and control actions over a given time period to determine the optimal control action for the current time. While this method tends to be computationally expensive, it provides optimal control actions. Due to its ability to account for many complex constraints, MPC is often considered for use in space applications despite the computational cost \citep{BashnickMPC, CairanoMPC, GoodyearMPC}.

Control Lyapunov Functions (CLF) and Control Barrier Functions (CBF) are functions designed to ensure system stability and safety, respectively. These functions are typically incorporated as constraints within an optimization framework, often with a quadratic cost function to create a controller. The dynamic model must be known as these functions are derived from system dynamics. However, this allows for the inclusion of many task-specific constraints into the controller design. CBFs have previously been used alongside reinforcement learning-based guidance algorithms, derived from the Clohessy-Wiltshire (CW) equations \citep{HibbardCBF, DunlapControls, DunlapRL}. In the discussion to follow, we focus on CBFs for safe reinforcement learning, derived using the CW equations. Consider a quadratic cost function of the following form, where $u_{ref}$ is the signal provided by the guidance algorithm:
\begin{equation}
    \min_u \ (u-u_{ref})^TQ(u-u_{ref})
\end{equation}

CBFs are applied as constraints to this equation to account for the safety and sensor requirements of the system. For ease of application, the agent is assumed to be fully actuated, with equal thrust capabilities along the principal axes in both the positive and negative directions and a minimum of 3 reaction wheels for full attitude control.

\subsubsection{Constraints}

To ensure a safe rendezvous and docking mission, agents must satisfy several operational constraints. First, the translational and angular accelerations are limited by the physical capabilities of the provided hardware. This limitation applies to each translational direction, $O=\{x,y,z\}$, and is formulated as:
\begin{equation}
    -F_{max, O}\leq u_{O}\leq F_{max,O}
\end{equation}
Similarly, the angular acceleration is limited in each direction, $R=\{\phi,\theta,\psi\}$, by:
\begin{equation}
    -T_{max, R}\leq u_{R}\leq T_{max,R}
\end{equation}

Following the formulation in \cite{HibbardCBF}, additional safety constraints are introduced. To prevent collisions, an agent must never occupy the same space as either the target or any other agent. Consider a set of S agents $i=1,...,S$ and $i\neq j$, each with some relative position vector $\vec\rho$ with respect to the target. Each agent and the target also have collision avoidance spheres with some radius $r$. With this information, the collision avoidance constraints are written as follows, where $\phi_1$ defines agent-agent interactions and $\phi_2$ defines agent-target interactions:
\begin{equation}
    \varphi_1 = ||\vec\rho_i-\vec\rho_j||_2-(r_i+r_j)\geq0
\end{equation}
\begin{equation}
    \varphi_2 =||\vec\rho_i||_2-(r_i+r_t)\geq 0
\end{equation}
The agent must also not drift so far from the target that the dynamic model representing the relative motion is no longer valid, defined by $r_m$. This value should be relatively small compared to the orbit radius. To ensure that drift is kept to a minimum, the following constraint is formulated:
\begin{equation}
    \varphi_3 = r_m-||\vec\rho_i||_2\geq 0
\end{equation}
The agent must also adhere to velocity restrictions. As the agent approaches the target, the agent's maximum allowable velocity should decrease to improve the chances of a collision-free docking mission. A safe relative velocity constraint that lowers the limit based on distance can be written as:
\begin{equation}
    \varphi_4=\nu_1||\vec\rho_i||_2+\nu_0-||\dot{\vec\rho_{i}}||_2\geq 0
\end{equation}
Finally, due to the line-of-sight (LOS) requirements of onboard sensors, the agent must avoid configurations where the Sun enters the camera’s FOV to protect the sensor from damage. Consider $\theta_s$ defining a conic exclusion zone and $\hat e_s$ as a time-varying unit vector pointing from the Sun to the target.
\begin{equation}
    \varphi_5 = -\frac{\langle \vec\rho_i,\hat e_s \rangle}{||\vec\rho_i||_2}+\cos\frac{\theta_s}{2}\geq0
\end{equation}

\subsubsection{Clohessy-Wiltshire Control Barrier Functions}

Based on the constraints outlined above, CBFs are derived using the CW equations. To formulate CBFs that contain all possible dynamic scenarios, \cite{HibbardCBF} defines a term $a_m$ describing the worst possible acceleration of the satellite, given the natural dynamics and the available control on the satellite:
\begin{equation}
    a_m = \frac{F_{max}}{m}-3n^2r_m-2nv_{m}
\end{equation}
where $v_m$ is the maximum allowable velocity and $n$ is the mean motion of the target's orbit. With this acceleration definition, constraints $\varphi_{1-5}$ correspond to the following CBFs $h_{1-5}$:
\begin{equation}
    h_{1,ij}=\sqrt{2(a_{m,i}+a_{m,j})(||\Delta\vec{\rho}_{ij}||_2-(r_i+r_j))}+\frac{\Delta\vec\rho_{ij}^T \Delta\dot{\vec{\rho}}_{ij}}{||\Delta\vec\rho_{ij}||}\geq 0
\end{equation}
\begin{equation}
    h_2=\sqrt{a_m(||\vec\rho_{i}||_2-(r_i+r_t))}+\frac{\vec\rho_{i}^T\dot{\vec\rho_{i}}}{||\vec\rho_{i}||}\geq 0
\end{equation}
\begin{equation}
    h_3 = \sqrt{a_m(r_m-||\vec\rho_i||_2)}+\frac{\vec\rho_i^T\dot{\vec\rho_i}}{||\vec\rho_i||}\geq 0
\end{equation}
\begin{equation}
    h_4=\nu_1||\vec\rho_i||_2+\nu_0-||\dot{\vec\rho_{i}}||_2\geq 0
\end{equation}
\begin{equation}
    h_5=\sqrt{a_m||\vec\rho_{i}-\vec\rho_{pr,i}||_2}+\frac{\vec\rho_i^T(\dot{\vec\rho}_i-\dot{\vec\rho}_{pr,i})}{||\vec\rho_i||}\geq 0
\end{equation}
More details on the derivations of these CBFs can be found in \cite{HibbardCBF}.

\subsection{Navigation}

The goal of the navigation system is to provide sufficient and accurate information about the environment for the guidance algorithm to use for maneuvering the satellite towards a desired state. In the case of rendezvous and proximity operations, this goal requires knowing where the target is with respect to the agent. However, in the known but uncooperative case of rendezvous, the agent cannot assume that the target will be able to communicate with other satellites. Therefore, the agent must be capable of measuring the target's state information independently, without relying on devices onboard the target. Monocular navigation has become a popular solution for obtaining target location in rendezvous in recent years due to its low size, weight, power requirements and cost (SWaP-C), making it ideal for small satellite applications.

\subsubsection{Feature Extraction Algorithms}

Processing images taken during flight can be done through traditional feature extraction algorithms such as SIFT \citep{SIFT} and SURF \citep{SURF}. Extracted features are combined with a known 3D model to determine the pose of the target. The major downside of these methods is the requirement of a 3D model of the target, which is often not available before flight. However, these methods are lightweight and easily implementable, therefore making them popular in many computer vision applications \citep{DLvsCV}.

\subsubsection{Convolutional Neural Networks} \label{sec:CNN}

There has been an increase in the use of convolutional neural networks (CNNs) to replace traditional feature extraction algorithms due to their versatility and increased accuracy \citep{DLvsCV}. CNNs are particularly popular for image processing due to their superior feature extraction abilities compared to other types of neural networks \citep{CNNoverview}. To accurately predict relative position and orientation, or pose, most networks used in rendezvous are split into three sections: the "backbone", the "neck", and the task-specific "heads". The backbone is the first section of a model, which is responsible for extracting general features from an image. This can include edges, corners, or other notable geometric or color characteristics. For example, both AlexNet \citep{AlexNet} and EfficientNet \citep{EfficientNet} are convolutional backbones that can provide the feature extraction necessary for more in-depth tasks. The neck portion of a model attaches the backbone to the heads, often enhancing the extracted features. Region proposal networks \citep{FasterR-CNN} are commonly used to propose regions of interest where features are being detected in an image. Feature pyramid networks like FPN \citep{FPN} and BiFPN \citep{EfficientDet} are also common object necks due to their ability to fuse features across different resolutions. The neck provides this additional information that can then be utilized by the heads. Heads are any tasks performed that are specific to the application that utilize the previously extracted feature information. Some of these tasks include bounding box regression, position and orientation (or pose) estimation, keypoint detection, heatmap generation, and semantic or instance segmentation. The pose head is an essential task for finding the position and orientation of a target without communicating with it, and will inform the guidance system about the environment. While other tasks are optional, they are sometimes included to increase robustness of a network and prevent overfitting to a specific task at the cost of increased model size and slower inference times \citep{Deep-6DPose, SPNv2}.

\subsubsection{Training Data Generation}

Neural networks require a large amount of training data to sufficiently learn the desired target domain. Due to the lack of available real images of satellites for neural network training, data must be made and collected through other means. However, care must be taken to ensure that they are as close to real images as possible, so that the models trained on them remain valid. This is mainly done through two types of images, those being synthetic and simulated images. Synthetic images are labeled images created in a simulation software using a 3D model of the target. Previous works have used Blender \citep{BlenderAura} or Unreal Engine \citep{Proenca2020} for image and label generation. Synthetic images are common since it is easy to generate large quantities of them, but they are often not as good at replicating the real space environment since a computer environment would invariably miss more obscure lighting effects. Simulated images, on the other hand, are images created in a physical laboratory setting, with scaled down models and lighting setups made to simulate real lighting effects that occur in space. These images tend to resemble actual on-orbit imagery more closely but are difficult to produce at scale due to the resource-intensive laboratory setup and the need for accurate labeling. The laboratory setup required for high-quality simulated image generation is discussed in Section \ref{sec:hil}. Since the generation of simulated images is more difficult, a combination of both types is used to create a large and robust dataset.

\subsubsection{Bridging the Space Domain Gap}

Despite efforts to create large, robust datasets on the ground, there is still a non-negligible difference between synthetic, simulated, and real images. This difference could cause the navigation network to produce less accurate results during deployment, potentially leading to collisions. One way to bridge the gap between synthetic and real images is to introduce data augmentations to the synthetic image set. Data augmentations are alterations applied to the synthetic images that cause them to have features closer to the desired testing domain. Random brightness, contrast, Gauss or ISO noise, and sun flares have all been previously used to augment synthetic images such that they contain image qualities more similar to the test domain \citep{KevinBlack, SPNv2, PVSAR}.

Another way to bridge the gap is to perform online or self training. Online training is an overarching term that refers to methods of training on images taken during a mission, after the initial supervised training period has occurred. This type of training is helpful in scenarios where real data is scarcely available or simulated/synthetic data generation does not accurately recreate the real environment that the network will be used in, despite data augmentation efforts. Since the space environment struggles with both of these issues, networks including online training methods can adapt to environmental changes easier than networks that do not include online training during a real-life mission. One common approach to online training is pseudo-labeling \citep{MKPNet, PVSAR}, where the network generates predicted labels, or pseudo-labels, for unlabeled images collected during the mission. These pseudo-labeled samples are then used to further fine-tune the network, allowing it to adapt to the real space environment without requiring manually labeled data. Another approach is Shannon entropy minimization \citep{TENT, SPNv2}. Entropy minimization increases model confidence and reduces error, allowing for the model to better account for discrepancies between training images and real images during deployment.

SPEED+ is a dataset created by the Stanford Space Rendezvous Laboratory (SLAB), which contains both synthetic and simulated images of the Tango satellite to test a model's ability to adapt in the presence of a domain gap \citep{SPEED+dataset}. This dataset splits the synthetic images into training and validation sets, and the simulated images are used as test sets. Without the availability of real images, the synthetic to simulated gap acts as a replacement for the actual domain gap. Networks often participate or compare themselves to the Satellite Pose Estimation Competition 2021 (SPEC), which tested a variety of models on the SPEED+ dataset \cite{SPECcomp}.

\subsubsection{Networks in Rendezvous}

SPN \citep{SPN} and SPNv2 \citep{SPNv2} are two models developed by SLAB. SPN was designed to take a monocular grayscale image as an input and solve for the 2D bounding box and pose of the target. SPN starts by feeding a grayscale image through five convolutional layers. This output is then given to three tasks: 2D bounding box classification and regression, relative attitude classification, and relative attitude regression. The bounding box information is solved through the use of a region proposal network, or RPN \citep{FasterR-CNN}. The attitude task combines the convolutional output and the bounding box output to determine the nearest predefined attitude class that satisfies the previously provided information. Then, using the Gauss-Newton algorithm, the pose is solved from the bounding box and attitude class given by the network. 

SPNv2 differs largely in comparison to SPN, and follows a more traditional outline, like discussed in Section \ref{sec:CNN}. The backbone of SPNv2 is modeled heavily after EfficientNet \citep{EfficientNet}. The neck, BiFPN \citep{EfficientDet}, acts as a feature fusion network that combines multi-scale information from the backbone. Ultimately, these features are received by three different heads: pose estimation, heatmap prediction, and segmentation. In contrast to SPN, the first head of SPNv2 uses EfficientPose \citep{EfficientPose} to perform object classification, bounding box prediction, and direct 6D pose regression without the use of predefined attitude classes. The second head, heatmap prediction, provides 2D heatmaps for the predicted locations of predetermined keypoints on the target. These keypoints can be used to solved the Perspective-n-Point problem \citep{EPnP}, providing an additional pose estimate. The last task, segmentation, separates the foreground from the background, allowing for pixel-level identification of the entire object. While not all of these tasks are required to deduce the desired navigation information, having a network train on multiple similar tasks reduces the chance of overfitting and increases its ability to generalize because multi-task training forces the network to learn shared features rather than task-specific features. This generalization helps the network be more adaptable to variations in future images, meaning that it will not experience as much of a decrease in accuracy during flight due to the domain gap. The tradeoff of including more tasks is that the network becomes larger and therefore typically takes longer to train and longer to produce results. SPNv2 also includes an online training method called online domain refinement (ODR). ODR updates the model by minimizing the Shannon entropy of the segmentation output only, rather than optimizing the full multi-task loss used during offline training, following a modified version of \cite{TENT}. This self-supervised refinement improves the network’s confidence in its predictions during the deployment phase. One restriction of ODR is that it updates only the normalization layers within the backbone and BiFPN to prevent model forgetting.

The network used at the SnT Zero-G Lab solves for pose from a single grayscale image \citep{OnGroundVivek}. This network is inspired by the Mask-RCNN structure \citep{Mask-RCNN}. The network uses HRNet \citep{HRNet} as the backbone, followed by a Region Proposal Network (RPN) \citep{Mask-RCNN} that identifies candidate regions of interest in the feature space. These regions are then processed by two downstream task heads: a keypoint prediction head and a bounding box regression head. The pose starts with the prediction of 2D keypoints using convolutional layers. Then, these keypoints are used to solve Perspective-n-Point (PnP) and produce a 6D pose vector. AISAT \citep{AISAT} uses a pose prediction structure similar to the SnT Zero-G Lab, finding pose through keypoint prediction and PnP. AISAT and Zero-G Lab networks have fewer "non-essential" tasks, allowing them to produce results faster than larger multi-task networks that predict outputs beyond the 6D pose.

\cite{KevinBlack} proposes a lightweight CNN for rendezvous missions. The backbone used in this network is MobileNetv2 \citep{MobileNetv2}, which was designed to run on mobile devices, making it computationally inexpensive in comparison to previously discussed networks. The backbone connects to a keypoint regression head whose outputs are used in EPnP to solve for 6D pose. With only 6.9 million parameters, this network fits well into CubeSats where computational resources are limited. Despite its size, this network's results were comparable to top submissions for the older SPEC competition which used the SPEED dataset \citep{SPEC_speed}.

\subsection{Hardware-in-the-Loop} \label{sec:hil}


Thorough ground testing of a GNC algorithm is a necessary step towards proving flight-readiness. Developing an on-ground laboratory that accurately represents the space environment will ensure that the GNC algorithm is being tested in an environment most similar to what would be experienced in space. Existing facilities often use 6-DOF robotic arms to simulate rendezvous trajectories and create laboratory environments that mimic space to demonstrate viability of GNC methods. Besides GNC testing, the laboratory can also be used for simulated data generation. 

One such facility is the Robotic Testbed for Rendezvous and Optical Navigation (TRON), which is managed by SLAB \citep{TRONcalibration}. This facility features two robotic arms that alter the relative pose between the agent and target by moving the two objects. The relative pose between the two objects is determined using either Vicon motion-tracking cameras or a combination of Vicon and KUKA robot telemetry, depending on the dataset generation setup. The latter configuration employs a multi-source calibration process that fuses both measurement systems to achieve millimeter-level position and millidegree-level orientation accuracy. The facility also recreates realistic space illumination conditions using lightboxes to simulate diffuse Earth albedo and a sunlamp to simulate direct sunlight, while blackout curtains provide the proper optical background. These capabilities enable the generation of hardware-in-the-loop (HIL) or simulated images (i.e., real photographs captured under controlled, space-like conditions). These accurately labeled simulated images are used alongside synthetic computer-rendered images to create datasets such as SPEED \citep{SPEEDdataset}, SPEED+ \citep{SPEED+dataset}, and SHIRT \citep{SHIRTdataset}.

The European Proximity Operations Simulator 2.0 (EPOS) located at the German Space Operations Center (GSOC) is another facility designed for rendezvous and docking simulations \citep{EPOSlab}. The facility features two six-degree-of-freedom KUKA robotic arms, one of which is mounted on a 25 m linear rail, allowing simulation of the relative motion between the agent and target. EPOS enables hardware-in-the-loop testing of sensors, docking mechanisms, and guidance, navigation, and control (GNC) systems over separations ranging from 25 m to contact. EPOS has cameras that are capable of capturing RGB images as well as depth images from the agent's perspective, allowing for a larger variety of available data. Alongside blackout curtains, a unique feature of this laboratory is that it can project images of Earth onto the curtains behind the target. Creating an imaging environment with Earth in the background helps test the robustness of the image-processing algorithms to variations in scene backgrounds. GSOC also conducted a review of sunlamps, focusing on matching solar irradiance and color temperature to produce a valid imaging environment.

Located at the University of Luxembourg, the SnT Zero-G Lab performs simulations of on-orbit servicing missions \citep{OnGroundVivek}. The two 6-DOF robotic arms sit on sliding rails, and the poses of the satellite mockups (rigidly attached to the end-effectors) are tracked via OptiTrack cameras for accurate ground-truth information during real-time simulations. \cite{LessonsLab} conducted an extensive market survey and experimental analysis of cameras, lamps, and background materials to select resources that would reflect the desired environment. This thorough analysis provided the Zero-G Lab with an accurate imaging environment for future experiments.

The Orbital Robotic Interaction, On-orbit servicing, and Navigation (ORION) laboratory located at the Florida Institute of Technology (FIT) features a unique robotic system compared to the previous experimental set ups covered in this section \citep{ORIONlab}. The agent and target each sit on pan-tilt mechanisms providing two rotational degrees of freedom (elevation and azimuth) for attitude control. The agent’s pan-tilt mechanism is mounted on a 2-DOF motion table, enabling translational motion in the horizontal plane and creating a 6-DOF relative-motion system between the chaser and target spacecraft. Surrounding the system is an epoxy air-bearing flat floor that allows free-floating targets for contact-dynamics experiments. Similar to other facilities, the ORION laboratory is painted black and features a high-intensity, dimmable LED sunlamp for replicating the imaging environment.

The Spacecraft Proximity Operations Testbed (SPOT), located at Carleton University, tests reinforcement learning-based guidance algorithms and navigation neural networks \citep{SPOTlab}. The facility includes three air-bearing spacecraft platforms capable of three-degree-of-freedom motion $(x,y,\theta)$. The target is outfitted with a docking port. The agent contains a variety of rendezvous sensors, including stereovision cameras, infrared cameras, and LiDAR systems. A third RSO is placed in the system to act as an obstacle for the agent to navigate around. Similar to navigation-only laboratories, a set of motion-tracking cameras provide accurate pose readings of the spacecraft to provide ground-truth information. Although a visual navigation system is used in this laboratory, the model satellites are not made to be visually similar to real satellites and there is no discussion of simulating the imaging environment.

\section{Problem Statement}\label{sec:ps}
There are multiple formulations of the docking problem that can be solved via the solution methods in this paper. The guidance results in this paper are produced from solving the single-agent docking problem, though future work will employ multi-agent reinforcement learning to solve docking problems involving multiple chasers. There are additional assumptions that may be invoked in problem formulation such as the introduction of velocity constraints and the selection of initial separation between the chaser(s) and the target. This is driven by the environment complexity and the need to establish a working solution on simpler problems before applying the solution methods to more complex docking scenarios.
\subsection{Governing Dynamics}\label{govdyn}
Consider a chaser spacecraft \(\textbf{C}_1\) in LEO that seeks to dock with a target spacecraft \(\textbf{T}_1\) that is also in the same orbit in LEO. The relative translational dynamics of these spacecraft can be formulated via the nonlinear Battin-Giorgi approach with a non-Keplerian reference orbit which is typically written in the Local Vertical Local Horizontal (LVLH) frame instead of an inertial frame. To define the relative translational dynamics, first consider an LVLH frame fixed at the center of the target spacecraft \(\textbf{T}_1\). Let the inertial position and velocity of the target spacecraft with respect to the central body be defined as \(\vec{r}_T \text{ and } \vec{v}_T\) respectively. The basis vectors of the LVLH frame are defined as follows:
\begin{itemize}
    \item The radial unit vector $\hat{r}=\frac{\vec{r}_T}{||\vec{r}_T||}$ which is parallel to the position vector of the target spacecraft with respect to the main body
    \item The angular momentum unit vector $\hat{h}=\frac{\vec{r}_T\times\vec{v}_T}{||\vec{r}_T\times\vec{v}_T||}$ which is parallel to the angular momentum of the target spacecraft
    \item The third unit vector $\hat{\theta}$ which completes the right-hand rule via $\hat{\theta}=\hat{h}\times\hat{r}$
\end{itemize}

\begin{figure} [!h]
    \centering
    \includegraphics[width=0.75\linewidth]{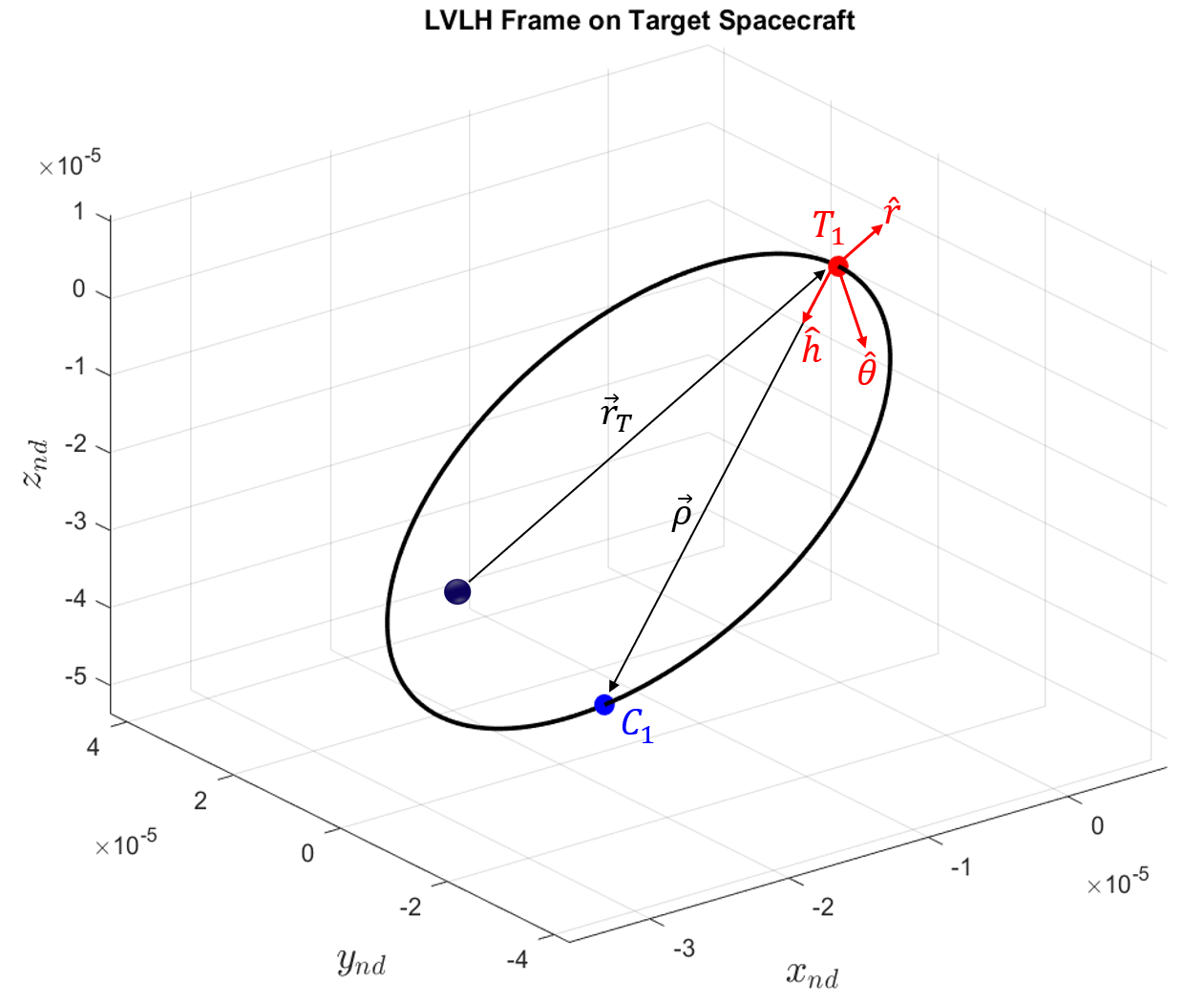}
    \caption{Chaser and Target Spacecraft in LVLH Frame, Figure Adapted from \cite{khourythesis2020}}
    \label{fig:lvlhframe}
\end{figure}

The target \(\textbf{T}_1\) and chaser \(\textbf{C}_1\) defined in the LVLH frame are visualized in Fig. \ref{fig:lvlhframe}. Considering the derivation from \cite{orbitalrendezvous_HPC} that is also formulated in \ref{app:rotdynamics}, let \(\mu\) represent the gravitational parameter of the main body, $\vec{a}_{T_p}$ and $\vec{a}_{C_p}$ are the perturbing accelerations on the target/chaser, and \(\vec{a}_{C_t}\) is the thrust acceleration of the chaser. Further, the transport theorem must be used to account for the rotation of the LVLH frame, which brings: 
\begin{align}
    ^N\dot{\vec\rho}=^R\dot{\vec\rho}+^N\vec\omega^R\times\vec{\rho}
\end{align}
where the time derivatives of relative position in inertial and rotating LVLH frames are $^N\dot{\vec\rho}$ and $^R\dot{\vec\rho}$, respectively. Letting the relative position between the chaser and target spacecraft be defined as \(\vec{\rho}=\vec{r}_C-\vec{r}_T\) (and subsequently \(\ddot{\vec{\rho}}=\ddot{\vec{r}}_c-\ddot{\vec{r}}_t\)), the complete equation for relative translational acceleration is provided below: 
\begin{align}
^R\ddot{\vec{\rho}} = -2^N\vec{\omega}^R\times ^R\dot{\vec{\rho}}\ -\ ^N\dot{\vec{\omega}}^R \times \vec{\rho} -\ ^N\vec{\omega}^R \times\ ^N\vec{\omega}^R \times ^R\vec{\rho} + \frac{\mu}{r_t^3} \frac{q\left[2+q+(1+q)^{1/2}\right]}{(1+q)^{3/2}\left[(1+q)^{1/2}+1\right]} \vec{r}_T- \frac{\mu}{r_c^3}\vec{\rho} + \vec{a}_{C_p} + \vec{a}_{C_t} - \vec{a}_{T_p}\end{align}
where \begin{align}
    q=\frac{\rho^2+2\vec{\rho}\cdot \vec{r_t}}{r_t^2}
\end{align}
\indent This formulation accounts for perturbing forces on the target and chaser, invokes no assumptions on relative distance, and assumes the target has no acceleration due to thrusting but the chaser has an acceleration due to thrust burning. Next, consider the following two spacecraft body frames: $F_{bs}$ is the body frame of spacecraft $S$ where the single-agent formulation considers $S=\textbf{C}_1$ to denote the single chaser spacecraft that has an inertial angular velocity $\vec{\omega}_{bs}$, and $F_{bt}$ is the body frame of the target spacecraft $\textbf{T}_1$ that has an inertial angular velocity $\vec{\omega}_{bt}$. Further, let $R_s, R_t$ denote rotation matrices that convert vectors from the frames $F_{bs}, F_{bt}$ to the inertial frame. The relative rotational dynamics are fully derived in \ref{app:rotdynamics} and the resulting transformation from $F_{bs}$ to $F_{bt}$ is provided below:
\begin{align}
    \ddot{\vec{\rho}}_{(bt)}=A\ddot{\vec{\rho}}_{(bs)}\ , \,\ A=R_t^TR_s
\end{align}
where the time derivative of the rotation matrix $A$ with respect to the inertial frame can be denoted as 
\begin{align}
    \dot{A}=-[\omega_{bt}^{(bt)}]_xA+A[\omega_{bs}^{(bs)}]_x
\end{align}
with the underscored $x$ denoting skew-symmetric matrices $[\ \cdot\ ]_x^T=-[\ \cdot\ ]_x^T$.

\subsection{Single-Agent Docking Problem}\label{subsec:singleagent}
Under the nonlinear Battin-Giorgi formulation, consider the target \(\textbf{T}_1\) and chaser \(\textbf{C}_1\) in the same Low-Earth Circular Orbit with the chaser starting at an initial distance \(d_i\) away from the target in the \(\hat{h}\) direction of the LVLH frame. The relative norm between the target and chaser must reach a magnitude less than the terminal position constraint represented by the tunable docking tolerance \(\epsilon\). The chaser \(\textbf{C}_1\) may perform impulsive thrusting and successfully docks if the magnitude of its relative position is less than the position tolerance \(\epsilon_p\). Safety constraints include avoiding collisions with other obstacles on the trajectory towards the target and contacting the target at a relative velocity whose magnitude is less than the maximum acceptable relative contact velocity \(\epsilon_{vel}\). The contact interface between the chaser and target shall either be a hardware-compatible docking port, or a derived pseudo-docking port as realized from future feature extraction methods. Following a successful dock, the angular momentum of the joint chaser-target system should be stable to prevent post-docking tumbling.
\subsection{Multi-Agent Extension}\label{subsec:multiagent}
A multi-agent extension to the docking problem introduces an additional chaser \(\textbf{C}_2\) that also seeks to dock with the target \(\textbf{T}_1\). The terminal criteria now requires both chasers to have docked, introducing a more rigorous collision avoidance requirement which requires the chasers to not collide with each other. This requires two separate (pseudo-)docking ports on the target spacecraft that can sufficiently be reached on collision-free trajectories. The simultaneous approach of multiple agents may require intermittent contacts to de-tumble the target and reduce its angular velocity prior to final capture. 
\section{Algorithms and Solutions}\label{sec:as}
The rendezvous and docking problem is solved through three major subsystems: guidance, navigation, and controls. State information is obtained from a dynamics-based state estimate combined with navigation measurements within a Kalman filter framework. The state estimate is then used as input to the guidance network to generate the guidance signal. This guidance signal is passed to the controller, which computes the necessary control inputs to drive the current state toward the guidance-commanded state while also ensuring safety. For real-time testing of a GNC solution, a laboratory space must be built to simulate the imaging and dynamic environment of space.
\subsection{Guidance}
The Deep Deterministic Policy Gradient (DDPG) algorithm is the preliminary method of accomplishing a guidance signal through RL \citep{ddpgoriginal}. This algorithm is used to solve the guidance problem for the single-agent case formulated in \ref{subsec:singleagent}, therefore all results presented in \ref{sec:guidanceresults} were produced from the DDPG algorithm on the single agent docking scenario. The DDPG variant that employs a distributional Bellman operator within a distributed learning framework is called Distributed Distributional DDPG (D4PG) \citep{Barth-Maron2018-d4pg}, which will be adapted and applied to the multi-agent problem in future configurations.
\subsubsection{DDPG}\label{subsubsec:ddpg}
The Deep Deterministic Policy Gradient (DDPG) algorithm is an actor-critic reinforcement learning framework designed for continuous action spaces. To enhance learning stability, DDPG utilizes target actor/critic networks that lag the original actor/critic networks through soft updates that are quantified by the coefficient \(\tau\). During interaction with the environment, transitions consisting of \((s_t,a_t,r_t,s_{t+1}, d_t)\) are stored in a replay buffer \(B\) where \(d_t\in\{0, 1\}\) is a flag that is set to 1 when the docking criteria is met. The critic samples random mini-batches of size $M$ to reduce temporal correlations, and these batches are used in computing target values and loss functions to optimize the actor/critic weights. \newline
\indent The algorithm forces exploration and avoids overfitting by using an epsilon-greedy explore-exploit strategy and noise added to each action. The exploration noise $dA$ is modeled as an Ornstein-Uhlenbeck (OU) process; given a random or neural network-informed action \(a_{t,original}\), the true action taken by the actor is:
\begin{align}
    a_t=a_{t,original}+dA\\ dA=\theta_{OU}(\mu_{OU}-a_t)dt_{OU}+\gamma_{OU}\sqrt{dt_{OU}}N(0,1)
\end{align}
where $\theta_{OU}=0.15$ is the rate of mean reversion, $\mu_{OU} = 0$ is the long-term mean, $dt_{OU} = .01$ is the time-step for discretizing the OU process, $\gamma_{OU}(t+1)=max(k\gamma_{OU}(t),.05)$ is a discount parameter to gradually reduce noise where $k=0.995$, and ${N}(0, 1)$ is a sample from a standard normal distribution. The RL algorithm is configured with the following state, action, and reward representations:
\begin{align*}
    \vec{s}_t=[\vec{x}_{t,target}-\vec{x}_{t,chaser},\vec{v}_{t,target}-\vec{v}_{t,chaser}]^T\\
    a_t=[a_{thrust}]\\
    r_{dense,t}=-c_1(||\vec{s}_t[1:3]||-||\vec{s}_{t-1}[1:3]||)\\
    r_{sparse,t}=R_{docked}, \text{\ if\ } ||\vec{s}_t[1:3]||<\epsilon_{pos}\ \&\ ||\vec{s}_t[4:6]||<\epsilon_{vel}\\
    r_{vel,t}=c_2\left(\frac{\epsilon_{vel}}{||\textbf{\textit{v}}_{t,target}-\textbf{\textit{v}}_{t,chaser}||}\right) \text{\ if\ } ||\vec{s}_t[1:3]||<3\epsilon_{pos}\\
r_{t}=r_{dense,t}+r_{sparse,t}+r_{vel,t}
\end{align*}
where the terms in the state vector \(\vec{x}_{t,(target, chaser)} \text{ and } \vec{v}_{t,(target, chaser)}\) represent the \(3\times1\) vectors for relative position and velocity between the target and chaser, the indexing notations $\vec{s}_t[1:3]$ and $\vec{s}_t[4:6]$ denote the subsets of $\vec{s}_t$ that only contain the position and velocity elements of the state vector respectively, and the action term \(a_{thrust}\) represents the acceleration due to thrust burning and may be parallel or anti-parallel to the direction towards the target ($\pm\hat{h}$ as visualized in Fig.\ref{fig:lvlhframe}). The terms in the reward function include the constant scalars \(c_1, c_2\) and the sparse reward for successful docking \(R_{docked}\) which are tunable parameters, the dense reward \(r_{dense,t}\) which is given to the agent continuously, the sparse docking reward which is only provided if the docking criteria is met, and the velocity reward which is provided when the chaser is within 3 position tolerances ($3\epsilon_p$) of the target. This velocity reward is a unique formulation because it incentivizes slower approaches rather than penalizing faster approaches. Additionally, the docking criteria may either require simultaneously satisfying the position and velocity criteria within \(\epsilon_{pos},\epsilon_{vel}\), or velocity constraints may temporarily be relaxed for the purpose of accomplishing intermediate solutions on simpler problems. Having defined the state, action, reward representation, the complete DDPG process is outlined in Algorithm 1. 
\begin{algorithm}[!h] 
\caption{Deep Deterministic Policy Gradient (DDPG)}
\begin{algorithmic}[1]\label{ddpg_alg}
\State Define learning hyperparameters: actor learning rate \(\alpha_0\), critic learning rate \(\beta_0\), target update rate \(\tau\), discount rate \(\gamma\), number of episodes \(N\), maximum steps \(T\) per episode, initial \& final exploration rates in epsilon-greedy \(\epsilon_0\ \&\ \epsilon_{min}\)
\State Define state and action sizes \(s_t \in \Re^{6}, a_t \in \Re^{1}\)
\State Instantiate actor network \(\pi_{\theta}\) and critic network \(Z_{\phi}\) with weights \(\theta, \phi\) respectively
\State Instantiate target actor network \(\pi'_{\theta'}\) and target critic network \(Z'_{\phi'}\) with weights \(\theta' \leftarrow \theta, \phi' \leftarrow \phi\)
\For{\(N\) episodes}
    \State Reset environment, $step=1$
    \While{\(docked = \text{False} \ \&\ step < T\)}
        \State \textbf{If}(\(rand(0, 1) < \epsilon_{\min} + (\epsilon_0 - \epsilon_{\min}) \cdot e^{-\lambda *step}):\  a_{t,original} = rand([-a_{\text{bound}}, a_{\text{bound}}])\)\ \textbf{Else} \(a_{t,original} = \pi_\theta(s_t)\)
        \State Actor performs an impulsive thrust burn with OU noise \(a_{t}=a_{t,original}+dA\) 
        \State Observe new state \(s_{t+1}\) \& reward \(r_{t+1}\), sum total rewards, (\(docked=\text{True}\) if docking \(tol\) met)
        \State Increment $step$
        \State Store tuple \((s_t, a_t, r_t, s_{t+1}, d_t)\) in replay buffer
        \State Critic randomly samples \(M\)-sized batch of tuples from buffer
        \State Compute TD target $y_t = r_t + \gamma (1 - d_t) Q_{\phi'}(s'_{t}, \mu_{\theta'}(s'_{t}))$
        \State Compute critic loss $L_c = \frac{1}{N} \sum_{t} (y_t - Q_{\phi}(s_t, a_t))^2$
        \State Compute actor loss $L_a = -\frac{1}{N} \sum_{t} Q_{\phi}(s_t, \mu_{\theta}(s_t))*(1-d_t)$
        \State Optimize actor/critic weights based on $\theta=\arg\min_{\theta}(L_a)$, $\phi=\arg\min_{\phi}(L_c)$
        \State Soft update target networks \(\theta' = (1 - \tau)\theta' + \tau\theta, \phi' = (1 - \tau)\phi' + \tau\phi\)
    \EndWhile
\EndFor
\State \textbf{return} optimal policy
\end{algorithmic}
\end{algorithm}

The two categories of algorithm development and tuning are manual tuning and automatic tuning. Because DDPG performance is highly sensitive to reward design and hyperparameters, both manual tuning and automatic tuning approaches are investigated.

\subsubsection{Manual Tuning}\label{sec412}
One possible method for tuning the learning algorithm involves manually shaping the reward function and other hyperparameters. The manual tuning process involves running a complete RL training, observing the performance via metrics such as the amount of successful docks and trends on final relative positions/velocities, and manually adjusting various hyperparameters in response. For example, if the agent continues to collide with the target at a velocity greater than the maximum allowable relative contact velocity, this observation would inspire the developer to introduce a greater collision penalty term in the reward function. This method requires substantial manual effort and is sensitive to small changes in the problem configuration, requiring ongoing tuning and retraining.

\subsubsection{Automatic Tuning}\label{alg_bayesianopt}
An alternative to manual tuning is automating the tuning process through hyperparameter tuning methods such as Bayesian optimization. The optimization algorithm first evaluates the objective function at several randomly selected hyperparameter settings to build an initial dataset, then constructs a surrogate model of the objective function. Using this surrogate, it takes well-informed steps by optimizing an acquisition function, resulting in rapid convergence that is favorable for optimizing expensive objective functions such as full RL training runs. \newline
\indent Each iteration of the Bayesian optimization runs an entire RL training over several thousand episodes to evaluate one hyperparameter configuration. The objective function is either quantified by success during training (such as maximizing the number of successful docks in the entire training or during a subset of episodes), or separate validation testing such as subjecting the agent to a test suite from various starting distances. The optimal designs identified from the Bayesian optimization are subsequently used in longer duration trainings with incrementally increasing complexity such as introducing additional constraints or increasing the initial distance \(d_i\).

\subsection{Optical Navigation Network}

\begin{figure}[h!bt]
    \centering
    \includegraphics[width=0.7\linewidth]{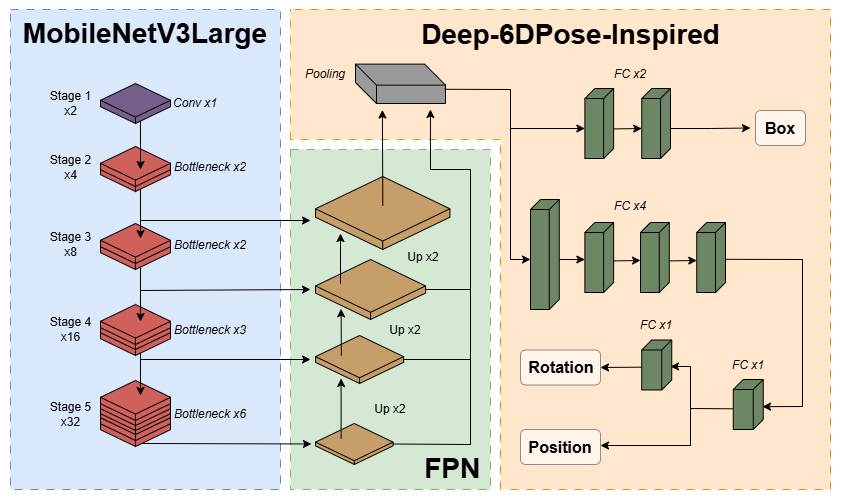}
    \caption{Overall CNN Architecture for Optical Navigation, Based on Combination of MobileNetV3Large, FPN, and Deep-6DPose \citep{MobileNetV3, FPN, Deep-6DPose}}
    \label{fig:DCNN}
\end{figure}

The proposed optical navigation network is a novel integration of MobileNetV3Large, FPN, and Deep-6DPose \citep{MobileNetV3, FPN, Deep-6DPose}. The full navigation network structure is shown in Fig. \ref{fig:DCNN}. 
MobileNetV3Large is a lightweight backbone, building off of previous MobileNet structures \citep{MobileNet, MobileNetv2}. These MobileNets prioritize creating small neural networks, suitable for deployment on mobile devices such as phones. MobileNetV3 is comprised of mostly bottleneck blocks, which are adapted from \cite{MobileNetv2}. MobileNetV3 adds squeeze and excitation to the residual layer inside the bottleneck and inserts swish activations to improve accuracy in comparison to previous versions. MobileNetV3Large is only 5.4M parameters, and 2.9M when excluding the ImageNet classification prediction layers which are not required for this application. This allows for flexibility in size when investigating potential architectures for the neck and heads.

Attached to the backbone is the Feature Pyramid Network (FPN) \citep{FPN}. The goal of the FPN is to collect features across multiple resolutions and fuse them to create a feature map containing both high and low level features. Without this portion of the network, the heads would only receive the output from the final stage of the backbone, with a very low resolution. With FPN, the heads can utilize information from any resolution and combine information across resolutions in order to produce more accurate results. In this application, FPN is attached to the last layer of stages 2, 3, 4, and 5. This combines 128x128, 64x64, 32x32, and 16x16 image resolutions. To perform this fusion, each stage output is upsampled to match the next stage size and then added together. For example, the stage 5 output is upsampled from 16x16 to 32x32, and then added to the stage 4 output. Then, the sum of stages 4 and 5 is upsampled again and added to the stage 3 output. This layer style of upsampling followed by addition is shown in orange in Fig. \ref{fig:DCNN}. This process is repeated until the final stage is reached. That singular output layer is then fed to the heads.

Deep-6DPose \citep{Deep-6DPose} is a multi-head network used for image processing, inspired heavily by Mask R-CNN \citep{Mask-RCNN}. Generally, training a multi-head network prevents overfitting to a specific task and allows a network to learn more general features, increasing its accuracy. The network described in Deep-6DPose has four outputs: object classification, bounding box, segmentation mask, and 6D pose. Due to the availability of labels in SPEED+ and information about the pictured satellite Tango, the presented work will only include bounding box and pose estimation tasks. The bounding box regression outputs the minimum and maximum pixel coordinates that result in the minimum area bounding box. The pose head is a direct estimator, meaning that PnP is not required to obtain the final pose. This task outputs a prediction for the translation vector $x,y,z$ and a Lie algebra \textit{so}(3) representation of the rotation. Exponential Rodrigues mapping \citep{Rodrigues} is applied to convert the predicted rotation from the Lie algebra \textit{so(3)} representation into the corresponding Lie group \textit{SO(3)} rotation matrix, which is then compared against the ground-truth rotation, typically expressed as quaternions or Euler angles. The MMEDR navigation network does not use the bounding box to calculate the $x,y$ as done in \cite{Deep-6DPose}, as this method saw slightly poorer performance than regular direct regression for this domain. Additionally, fully connected (or dense) layer sizes are reduced since the network is only learning the pose of one object and not multiple. Finally, the rotation task receives one supplemental dense layer since rotation is more difficult to learn than position in terms of direct regression.

\subsubsection{Data Augmentation} \label{sec:augments}

Data augmentation is an easy way to improve the applicability of synthetic images to the real imaging domain. Augmentations performed follow previous augmentation styles presented in a variety of works \citep{KevinBlack,SPNv2,SPECcomp}. This includes random brightness and contrast, Gauss or ISO noise, motion blur, and sun flares. Additionally, gamma augmentations have been introduced to simulate the overexposure that occurs in extremely bright lighting conditions. Image augmentations are performed using the Albumentations library \citep{Albumentations}. The augmentations are split into two domains: lightbox and sunlamp, inspired by the lightbox winners of SPEC 2021 \citep{SPECcomp}. Because of the extreme conditions and sun flares that can occur in the sunlamp domain, two different augmentation pipelines are made to allow for mildly augmented and heavily augmented images to be created. The lightbox and sunlamp augmentations share all the same augmentation types, except for sun flares which only occur in the sunlamp domain. The lightbox domain is given smaller ranges for the random application of augments, as it often has softer imaging conditions. Every image is augmented through either the lightbox or sunlamp augmentation steps, and each individual augment is applied with a certain probability, listed in Table \ref{tb:aug}. Gaussian noise, motion blur, and ISO noise are applied using Albumentations' OneOf command, picking one of the three types to apply. In addition to probability of application, the sunlamp augmentations are harsher, reflecting the difficulty of viewing under direct sunlight. Of all the images in the training set, 40\% go through the lightbox augmentation and 60\% receive the sunlamp augmentation. Although the test set is comprised of more lightbox images than sunlamp images, it is easier to learn the soft lightbox features. Therefore, emphasis is applied to learning scenarios with harsh lighting and occlusion by making the sunlamp augments more than 50\% of the total augments.

\begin{table}[h!bt]
    \centering
    \caption{List of Augmentations and Probability of Application}
    \begin{tabular}{c@{\qquad}ccc@{\qquad}ccc}
      \toprule
      Augment Type & Probability - Lightbox & Probability - Sunlamp \\
      \midrule
      Brightness and Contrast & 0.8 & 0.8  \\
      Gaussian Noise & 0.7 &  0.8 \\
      Motion Blur & 0.7 &  0.8 \\
      ISO Noise & 0.7 & 0.8 \\
      Gamma & 0.6 & 0.7 \\
      Sun Flare& - & 0.7\\
      \bottomrule
      \label{tb:aug}
    \end{tabular}
\end{table}

\subsubsection{Losses and Implementation Details}

The total loss across the position, orientation, and bounding box is found as:
\begin{equation}
\mathcal{L}_{total}=c_1\mathcal{L}_{t}+c_2\mathcal{L}_r+c_3\mathcal{L}_{box}
\end{equation}
where $c_1,c_2,c_3$ are scalar weights that determine how much each task's loss influences the overall loss. This allows for the emphasis of tasks deemed more important, such as pose estimation. The bounding box and rotation losses are defined by a smooth L1 function, similar to \cite{Mask-RCNN} and \cite{Deep-6DPose}. The translation loss performs better using an L2-norm. For this application, the weights are chosen to be $c_1 = 2,c_2=12,c_3=0.5$. $c_2$ is much larger because the angle loss is smooth L1, meaning the magnitude must be increased to match the other losses.

The navigation network is evaluated on the SPEED+ dataset \citep{SPEED+dataset} and compared to results reported in the SPEC 2021 competition \citep{SPECcomp}. The original image size provided in SPEED+ is 1920x1200. For application in the proposed model, images are resized using the original aspect ratio to 512x512 with padding. Although performing cropping instead of padding would maintain a higher resolution, there is a chance of cropping out large portions of the satellite, leading to unidentifiable images. The proposed model is built to handle RGB images, but has been adapted to accept grayscale inputs for testing on the SPEED+ dataset \citep{SPEED+dataset}. The navigation network uses MobileNetV3Large pretrained on the ImageNet dataset. The learning rate schedule is a cosine decay function which has an initial learning rate of 5e-4 and runs for 75 epochs with a linear warmup stage of 5 epochs.

The competition compares on three metrics, $E_t, E_q,$ and $E_p$, for position, orientation, and total pose error. Listed below are the equations used to calculate these metrics:
\begin{equation}
    \begin{split}
    e_t &= ||\hat{t}-t|| \\
    e_q &= 2\cos^{-1}(|\langle\hat{q}, q\rangle)\\
    e_p &= \frac{||\hat{t}-t||}{||t||}+2\cos^{-1}(|\langle\hat{q}, q\rangle)
    \label{eq:little_metrics}
\end{split}
\end{equation}

where $\hat{t}, \hat{q}$ are the ground truth position and orientation and $t, q$ are the predicted position and orientation. The Eq. \ref{eq:little_metrics} values are averaged over a batch size of $N$ to find $E_t, E_q,$ and $E_p$. Additionally, predictions whose accuracy beats the calibration values on the test set are set to zero. More details about the calibration threshold and the general calibration process in the TRON facility can be found in \cite{SPEED+dataset, TRONcalibration}.
 
\subsubsection{State Filter}

To update the state vector, propagated dynamics and measurements will be combined using a sigma-point–based Unscented Kalman Filter (UKF) once the MMEDR-Autonomous experimental setup is fully operational. This work adopts the framework proposed by \cite{DelayKalman}, which extends the Kalman filtering approach to handle delayed and asynchronous measurements. In our implementation, an unscented variant of this method will be employed to accommodate nonlinear dynamics while preserving the ability to handle non-constant delays. Details of UKF, which is described below in general terms, can be read in \cite{UKF}.

The UKF for an n-dimensional system starts with some initial mean and covariance of the state, $m_{k-1}^+=m_0$ and $P_{k-1}^+=P_0$. Then, the mean and covariance are used to form the sigma points $\mathcal{X}_{k-1}^{(i)}$ which are propagated forward to the next time step through the dynamic model:
\begin{equation}
    \begin{split}
        \mathcal{X}_{k-1}^{(0)} = m_{k-1}^+, \quad \mathcal{X}_{k-1}^{(i)} = m_{k-1}^++\sqrt{n+\lambda}\left[\sqrt{P_{k-1}^+}\right]_i, \quad \mathcal{X}_{k-1}^{(i+n)} = m_{k-1}^+-\sqrt{n+\lambda}\left[\sqrt{P_{k-1}^+}\right]_i \\
        w_0^{(m)}=\frac{\lambda}{n+\lambda}, \quad w_i^{(m)}=\frac{1}{2(n+\lambda)}, \quad  w_0^{(c)}=\frac{\lambda}{n+\lambda}+(1-\alpha^2+\beta), \quad w_i^{(c)}=\frac{1}{2(n+\lambda)}\\
        \mathcal{X}^{(i)}_k = f(\mathcal{X}_{k-1}^{(i)}), \,\ i=1,...,2n
    \end{split}
\end{equation}
where $\lambda = \alpha^2(n+\kappa)-n$ is the scaling parameter, $\alpha$ and $\kappa$ determine the spread of sigma points around the mean, and $\beta$ incorporates prior information. For example, if a distribution is known to be Gaussian, $\beta = 2$. If no information needs to be introduced, then $\beta$ can be set to zero. The predicted mean and covariance for the current time step can then be solved as:
\begin{equation}
    \begin{split}
        m_k^-=\sum_{i=0}^{2n}w_i^{(m)}\mathcal{X}_k^{(i)}\\
        P_k^- = \sum_{i=0}^{2n}w_i^{(c)}(\mathcal{X}_k^{(i)}-m_k^-)(\mathcal{X}_k^{(i)}-m_k^-)^T+Q_{k+1}
    \end{split}
\end{equation}
where $Q_{k+1}$ is some additive noise. In the event of non-additive noise, the noise can be augmented to the state and additional sigma points can be generated to propagate that noise. For simplicity, this work will cover the additive case. The sigma points will now be resolved, using the predicted mean and covariance, $m_k^-$ and $P_k^-$:

\begin{equation}
    \begin{split}
        \mathcal{X}_{k}^{-(0)} = m_{k}^-, \quad \mathcal{X}_{k}^{-(i)} = m_{k}^-+\sqrt{n+\lambda}\left[\sqrt{P_{k}^-}\right]_i, \quad \mathcal{X}_{k}^{-(i+n)} = m_{k}^--\sqrt{n+\lambda}\left[\sqrt{P_{k}^-}\right]_i \\
        w_0^{(m)}=\frac{\lambda}{n+\lambda}, \quad w_i^{(m)}=\frac{1}{2(n+\lambda)}, \quad  w_0^{(c)}=\frac{\lambda}{n+\lambda}+(1-\alpha^2+\beta), \quad w_i^{(c)}=\frac{1}{2(n+\lambda)}
    \end{split}
\end{equation}
These sigma points are then transformed through the measurement model to find the predicted measurement at time $t_k$. The predicted measurement can then be used to solve for the measurement covariance $W_k$ and the cross-covariance $C_k$:
\begin{equation}
\begin{split}
    \mathcal{Z}_k^{(i)}=h(\mathcal{X}_{k}^{-(i)})\\
    \hat z_k=\sum_{i=0}^{2n}w_i^{(m)}\mathcal{Z}_k^{(i)}\\
    W_k =\sum_{i=0}^{2n}w_i^{(c)}(\mathcal{Z}_k^{(i)}-\hat z_k)(\mathcal{Z}_k^{(i)}-\hat z_k)^T \\
    C_k = \sum_{i=0}^{2n}w_i^{(c)}(\mathcal{X}_k^{-(i)}-m_k^-)(\mathcal{Z}_k^{(i)}-\hat z_k)^T
\end{split}
\end{equation}
Finally, the Kalman gain can be computed, and the updated mean and covariance are found as:
\begin{equation}
    \begin{split}
        K_k = C_kW_k \\
        m_k^+ = m_k^-+K_k(z_k - \hat z_k) \\
        P_k^+=P_k^--C_kK_k^T-K_kC_k^T+K_kW_kK_k^T
    \end{split}
\end{equation}

Ideally, the full state estimate is a combination of the predicted state from dynamics and the measured state, assuming that there will be a measurement available at time $t_k$. However, due to a variety of delay sources, it is unlikely that there will be measurements available at each time step $t_k$. For example, the ML-based algorithms may take a variable amount of time to process an image or output the next ideal control action. In fact, the filter may need to perform multiple updates before the next measurement is available, depending on the $\Delta t$ between $t_{k-1}$ and $t_k$ and the total sum of system delays. Consider the scenario where a measurement has not been taken since the previous state update. In this case, only the dynamic propagation is used and the calculation of the predicted measurement is skipped, so the mean and covariance are set as $m_k^+=m_k^-$ and $P_k^+=P_k^-$. When a measurement becomes available, \cite{DelayKalman} discusses how to extrapolate a measurement given at some time $t_{meas}$, where $t_{meas}<t_k$, so that it can be used at the current time step. Adapted to UKF, this extrapolated measurement is written as:
\begin{equation}
    z_k^{extra}=z_{meas}+\sum_{i=0}^{2n}w_i^{(m)}h(\mathcal{X}_{k}^{-(i)})-\sum_{i=0}^{2n}w_i^{(m)}h(\mathcal{X}_{meas}^{-(i)})
    \label{eq:extra}
\end{equation}

The first two terms in Eq. \ref{eq:extra} are known at time $t_k$. However, since $t_{meas}$ does not occur on some count of $k$, the sigma points $\mathcal{X}_{meas}^{-(i)}$ are not immediately available. \cite{DelayKalman} presents two cases of measurement collection, which determines whether to use forward or backward propagation to solve for the sigma points at time $t_{meas}$ from the closest corrected state estimate (a state estimate updated with both the dynamics and measurements). In order to effectively implement forward propagation, some number of previous state estimates must be stored. Take $N$ to be the number of previously stored time steps such that state estimates from $t_{k-N},...,t_{k-1},t_k$ are available. $N$ should be chosen such that the time from $t_{k-N}$ to $t_k$ is larger than the combination of all potential system delays.

The above filter considers a single measurement from one camera only. Since these agents are multi-sensor, data must be combined in such a way that the resulting combination is often more accurate than the individual measurements. Shown in \cite{OWA}, Ordered Weighted Averaging (OWA) is an effective method of combining sensor data such that the weighted state estimate is closer to the true state than the individual state estimates. Consider three UKF results: UKF\textsubscript{1}, UKF\textsubscript{2}, and UKF\textsubscript{fusion}. UKF\textsubscript{1} and UKF\textsubscript{2} are the UKF results of analyzing cameras 1 and 2 separately. UKF\textsubscript{fusion} is an extension on the previously presented work, where the measurement vector length is extended such that it matches the total length of both sensor 1 and sensor 2's measurements. These three UKF results each have a state covariance matrix, which can be used to form weighting terms. Consider the first element in each of the UKF covariances $s_1,s_2, s_3$. Since values with a high covariance should be considered less in the overall estimate, the weights are structured as the inverse of the covariances. The weights must also be normalized, so the final weights are found as \citep{OWA}:
\begin{align}
    w_1 = \frac{\frac{1}{s_1}}{\frac{1}{s_1}+\frac{1}{s_2}+\frac{1}{s_3}}=\frac{s_2 s_3}{s_1 s_2 + s_2 s_3 + s_1 s_3} \\   
    w_2 = \frac{s_1 s_3}{s_1 s_2 + s_2 s_3 + s_1 s_3} \\
    w_3 = \frac{s_1 s_2}{s_1 s_2 + s_2 s_3 + s_1 s_3}  
\end{align}

This process is repeated for each covariance value in the matrix diagonal. The weights are then applied to their associated state estimate terms, and the terms are summed across the three weighted UKF results to produce one fused result, UKF\textsubscript{OWA}.

\section{Experimental Platform and System Integration}\label{sec:integration}

\begin{figure}[h!bt]
    \centering
    \includegraphics[width=0.75\linewidth]{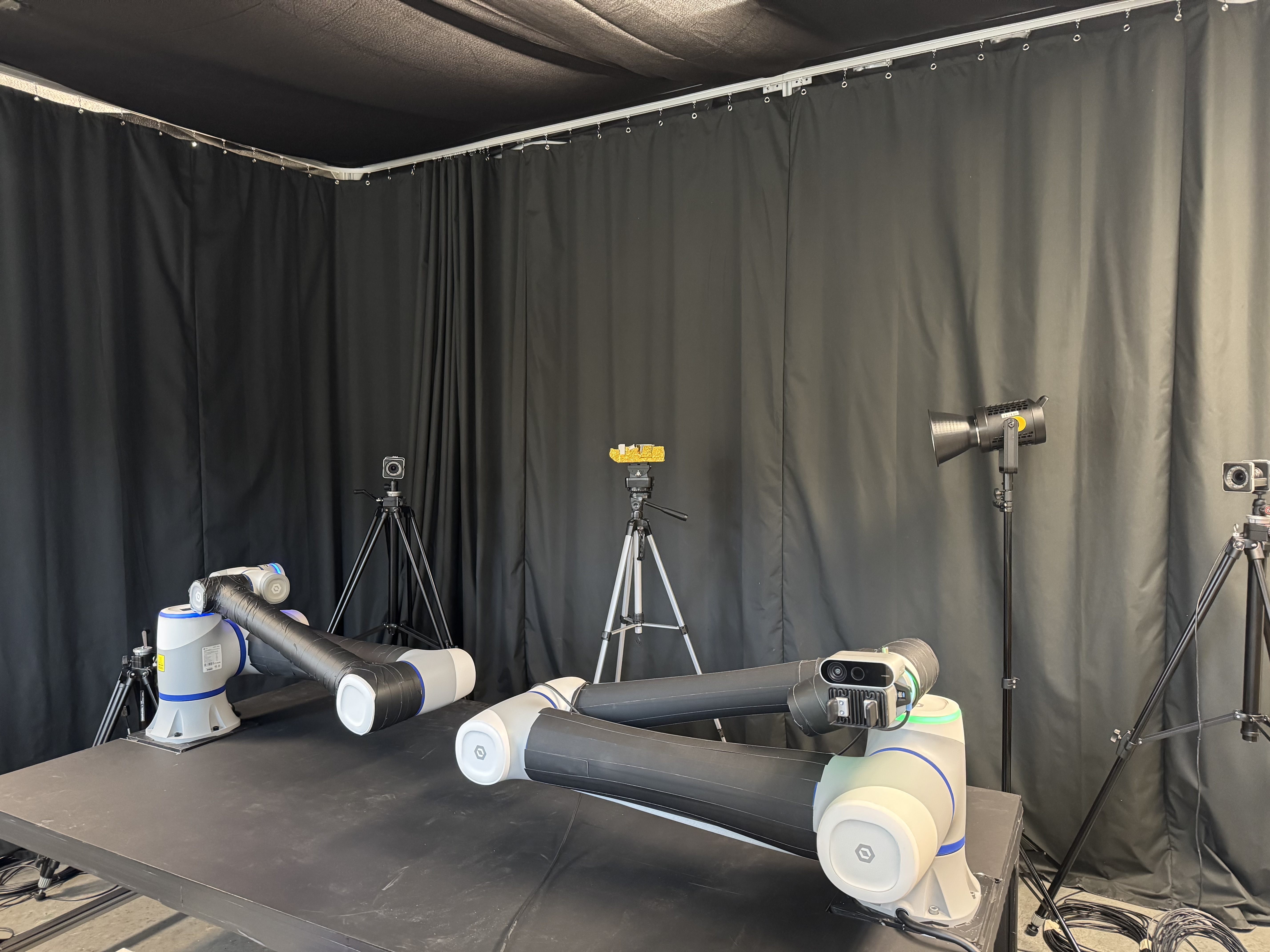}
    \caption{Current State of the DCV-Space Testing Facility}
    \label{fig:lab_photo}
\end{figure}

To perform experiments using multiple agents, the DCV-Space group is equipped with two 6-DOF CR20A robotic arms, with each arm carrying one agent. The Aura spacecraft was chosen as the target object due to its size. The large nature of this satellite necessitates the use of multiple agents for either servicing or de-orbiting. The target is a 3D-printed scaled model of the Aura satellite, which is placed on a pan-tilt camera stand. Being able to manually alter the target’s position and attitude using the pan–tilt stand, as well as adjust the agents’ poses via the robotic arms, enables the testing of a variety of initial target–agent relative poses between experiments. Stationed around the testbed are 8 Vicon Vero v2.2 cameras, used for producing accurate ground truth data for the agents and target. This information is vital for ensuring that the navigation network is producing reliable pose estimates. To test on realistic equipment, the on-board computer that will be running the GNC algorithm is an NVIDIA Jetson TX2. Each agent is fitted with two Orbbec FemtoBolt cameras. These cameras contain both RGB and depth imaging capabilities, allowing for the expansion into the depth realm for future research. Completing the imaging environment, blackout curtains cover the walls and windows to block out any external light, and a Godox UL60Bi lamp provides artificial sunlight. A picture of the current state of the laboratory space is shown in Fig. \ref{fig:lab_photo}. The final connections of software and hardware are still being made at this time.

To enable feasible experimentation, the distance and time of a typical rendezvous mission can be appropriately scaled to match the laboratory environment. Adopted from \cite{GoodyearMPC}, parameters $\kappa$ and $\nu$ are responsible for scaling the time and distance, respectively. These values are chosen based on the size of the laboratory environment and the desired experiment duration. The GNC algorithm first operates entirely in the "full-scale" orbital frame. Then, using $\nu/\kappa^2$, the resulting accelerations from the GNC algorithm are scaled into the laboratory frame. Once robotic motion has concluded and measurements are taken, those measurements are scaled back into the orbit frame for the process to repeat. As shown in Fig. \ref{fig:lab_photo}, the model of the Aura spacecraft is scaled down by this same factor so that the imaging environment stays equivalent in the laboratory and orbit frames.

\subsection{Robot Kinematics}
\begin{figure}[h]
    \centering
    \includegraphics[width=0.35\linewidth]{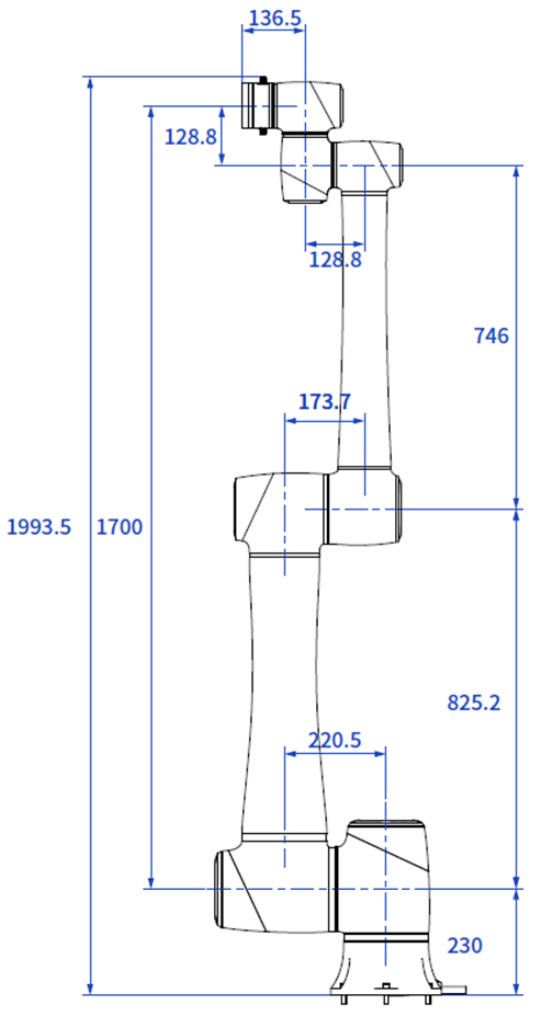}
    \caption{CR20A Link Dimensions}
    \label{fig:cr20adim}
\end{figure}

The robots utilized in the laboratory are the Dobot CR20A robotic arms. A diagram of the robot and dimensions are shown in Fig. \ref{fig:cr20adim}. These robots feature six Degrees Of Freedom (6-DOF), with joints 1-3 mainly affecting the position of the end-effector and joints 4-6 forming the "wrist" mostly determining the orientation of the end-effector, although some coupling exists. It is necessary to determine the joint angles to achieve a particular end-effector position and orientation. Like in \cite{OnGroundVivek}, this is accomplished using the inverse Jacobian method, which relates cartesian velocity to joint angular velocity in Eq. \ref{eq:veltransform}, where $\bold{\dot{x}}$ is the set of linear and angular velocities of the end-effector, $\bold{\dot{q}}$ is the set of joint angular rates, and $\bold{J}$ is the Jacobian matrix to transform angular rates to cartesian velocities.
\begin{equation} \label{eq:veltransform}
    \bold{\dot{x}}=\bold{J}\bold{\dot{q}}
\end{equation}
To solve for the joint angular velocity, the equation is multiplied by the Jacobian pseudo-inverse, resulting in:
\begin{equation}
    \bold{J^{+}}\bold{\dot{x}}=\bold{\dot{q}}
\end{equation}
where $\bold{J}^+$ represents the Jacobian pseudo-inverse given by:
\begin{equation}
    \bold{J^{+}}=(\bold{J}^T\bold{J})^{-1}\bold{J}^T
\end{equation}
Commonly, for 6-DOF robot arms, the position and orientation of the end-effector may be decoupled into joints 1-3 and 4-6 respectively, and the Jacobian may be found using simple geometry. However, the wrist for the CR20A is not spherical, resulting in coupled dynamics. To solve for the Jacobian, the Denavit–Hartenberg (D-H) parameters \citep{DHParams} are used to find the joint frame transformations. The table of D-H parameters for the CR20A Dobot are shown in \autoref{tb:dhparams}.
\begin{table}[h!bt]
\centering
    \caption{D-H Parameters for Dobot CR20A Robot}
    \label{tb:dhparams}
    \begin{tabular}{ |c|c|c|c|c| }
        \hline
        Link & $\theta$ & $d$ (mm) & $a$ (mm) & $\alpha$ (deg) \\ \hline
        1 & $\theta_1$ & 230 & 0 & 90 \\ \hline
        2 & $\theta_2$ & 0 & 825.2 & 0 \\ \hline
        3 & $\theta_3$ & 0 & 746 & 0 \\ \hline
        4 & $\theta_4$ & 175.6 & 0 & -90 \\ \hline
        5 & $\theta_5$ & 128.8 & 0 & 90 \\ \hline
        6 & $\theta_6$ & 136.5 & 0 & 0 \\ \hline
    \end{tabular}
\end{table}
\newline The transformation matrix from a previous link frame to the current link frame is found using the following equation \citep{DHParams}, which incorporates translational and orientation transformations:
\begin{equation}
    ^{i-1}\bold{T}_i=
    \begin{bmatrix}
        cos(\theta_i) & -sin(\theta_i)cos(\alpha_i) & sin(\theta_i)sin(\alpha_i) & a_i cos(\theta_i) \\
        sin(\theta_i) & cos(\theta_i)cos(\alpha_i) & -cos(\theta_i)sin(\alpha_i) & a_i sin(\theta_i) \\
        0 & sin(\alpha_i) & cos(\alpha_i) & d_i \\
        0 & 0 & 0 & 1
    \end{bmatrix}
\end{equation}
where, $\theta_i$, $\alpha_i$, $a_i$, and $d_i$ represent the joint angle, link twist, link length, and link offset, respectively. The complete transformation from the base frame (frame 0) to the 6th link frame (frame 6) can be expressed as:

\begin{equation}
        {}^{0}\bold{T}_{6}
    = {}^{0}\bold{T}_{1}\,
      {}^{1}\bold{T}_{2}\,
      {}^{2}\bold{T}_{3}\,
      {}^{3}\bold{T}_{4}\,
      {}^{4}\bold{T}_{5}\,
      {}^{5}\bold{T}_{6}
\end{equation}

Note that each of these transformation matrices are functions of the joint angles. The transform $^0\bold{T}_{6}$ can be expressed as a combination of a rotation matrix representing the orientation and the cartesian position vector of the end-effector in the base frame \citep{DHParams}:
\begin{equation}
    ^{0}\bold{T}_6=
    \begin{bmatrix}
        R_{11} & R_{12} & R_{13} & p_x \\
        R_{21} & R_{22} & R_{23} & p_y \\
        R_{31} & R_{32} & R_{33} & p_z \\
        0 & 0 & 0 & 1 \\
    \end{bmatrix}
\end{equation}

From the rotation matrix and position coordinates, the partial derivatives with respect to the joint angles can be computed to find the Jacobian. After the Jacobian is found, the pseudo-inverse can be computed. When the robot approaches singularities (areas of motion where the robot requires extremely large joint motions to produce small end-effector movements), the Jacobian becomes singular, and calculated joint velocities greatly exceed limits of the robot. Following \cite{OnGroundVivek}, this issue is resolved by scaling down the robot joint rotation rates by the condition number $\lambda$ of the Jacobian using the following formula:
\begin{equation}
\bold{\dot{q}}= \begin{cases}\bold{\dot{q}} & \lambda(\bold{J})\le\lambda_{l} \\
\bold{\dot{q}}\bigg(1-\frac{\lambda(\bold{J})-\lambda_{l}}{\lambda_{u}-\lambda_{l}}\bigg) & \lambda_{l}<\lambda(\bold{J})<\lambda_{u} \tag{15} \\
\bold{0} & \lambda(\bold{J})\ge\lambda_{u}\end{cases}
\end{equation}
where $\lambda_l$ and $\lambda_u$ are the lower and upper limits, respectively, for which damping on the joint velocity is required. These values are determined by laboratory-specific requirements.

\subsection{Lab Calibration and Data Fusion}

To accurately characterize the laboratory geometry, a calibration procedure is required. Performing laboratory calibration enables the generation of ground-truth pose labels by finding the unknown, fixed transformations between various frames \citep{TRONcalibration}. The laboratory contains two sources of measurements, one coming from the Vicon camera system and the other coming from the Dobot telemetry. These measurement sources are fused using the MAP estimate method to ensure the highest accuracy possible when generating pose labels. The laboratory calibration and data fusion procedure will be derived and executed when the laboratory is ready to run simulations.

\subsection{Integration}

\begin{figure}[!h]
    \centering
    \includegraphics[width=1\linewidth]{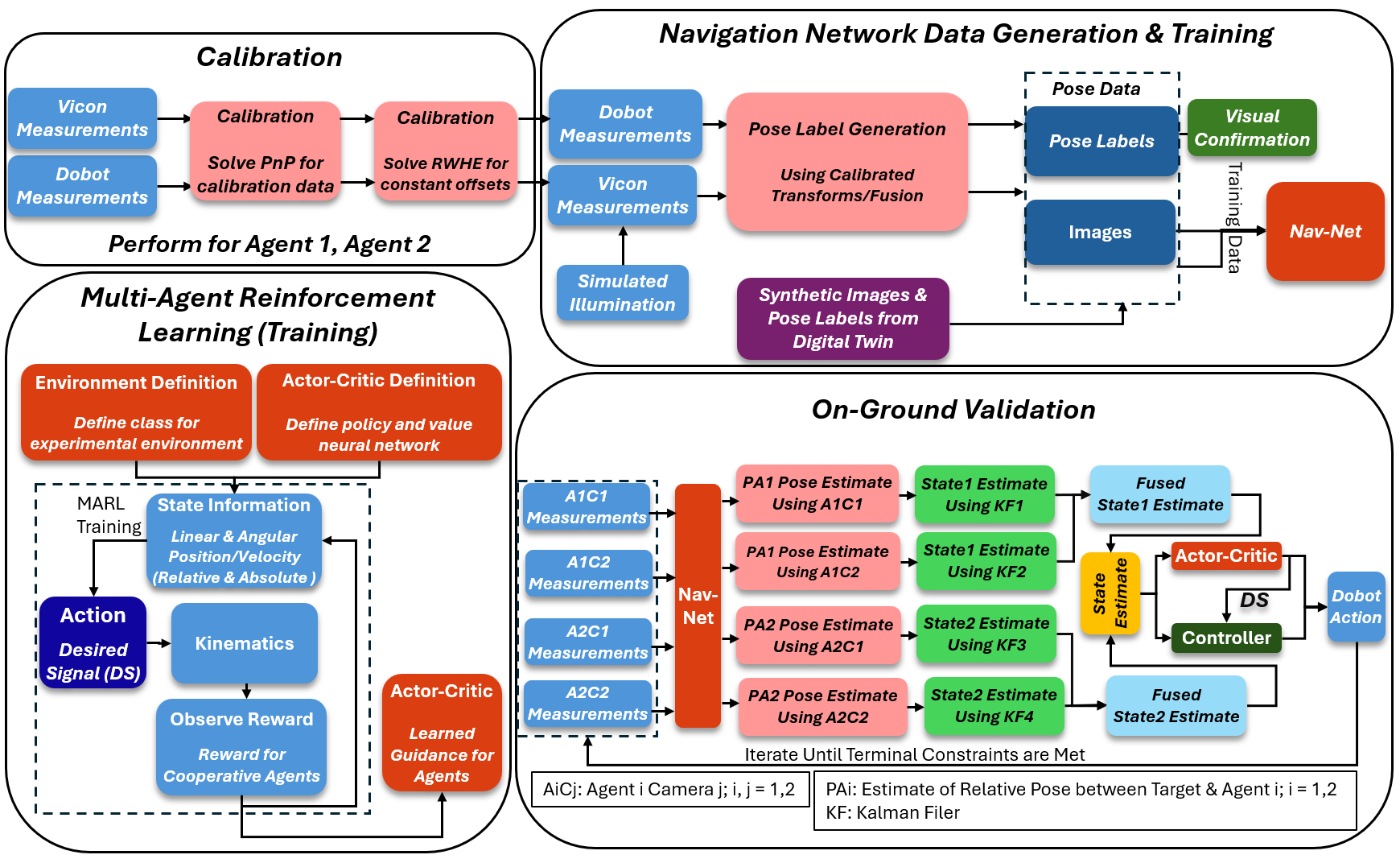}
    \caption{Overview of All MMEDR Subsystems}
    \label{fig:mmedr_integration}
\end{figure}

The collection of all MMEDR workflows is visually summarized in Fig.  \ref{fig:mmedr_integration}. During on-ground validation, the process starts with imaging the target spacecraft. Each of the two agents will acquire images using its two onboard cameras and transmit those images to its onboard computer. All four images will be processed by the optical navigation network to produce four pose estimates, two per agent. Each of these pose measurements will be fed into a Kalman filter to estimate the state, which consists of the absolute position, velocity, orientation, and angular velocity of the agent as well as the corresponding quantities for relative motion between the agent and the target. For each agent, the state estimates (from the two Kalman filters, each corresponding to an onboard camera as measurement source) will be fused together using ordered weighted averaging \citep{OWA} resulting in a single state estimate per agent. The states of the two agents are stacked together to form the overall state. The overall state subsequently informs the guidance algorithm, which returns an acceleration signal. Once the guidance network outputs the desired linear and angular accelerations, a constraint-aware controller computes the corresponding thrust and torque commands, ensuring convergence to the desired relative pose while satisfying system constraints via control barrier functions.

Once the actual linear and angular accelerations are available, the information is transferred to the laboratory computer, which controls the robotic arms. This computer will convert the acceleration in the orbit frame to an equivalent robotic command in the laboratory frame. The motion is enacted, and the process is repeated iteratively until the agents reached their desired relative positions and orientations.
\section{Preliminary Results}\label{sec:results}
\subsection{Guidance}\label{sec:guidanceresults}
As a preliminary step towards the multi-agent case in Section \ref{subsec:multiagent}, this section considers the single-agent rendezvous and docking problem formulated in Section \ref{subsec:singleagent}. A guidance policy is learned to map relative position and velocity states to thrust commands that achieve docking within a prescribed tolerance and acceptable contact velocity, with the RL state, action, and reward design defined in Section \ref{subsubsec:ddpg}. This section demonstrates results from learning a guidance policy via manual tuning and automatic tuning.
\subsubsection{Manual Tuning}\label{manualtuningresults}
\begin{figure}[h!]
    \centering
    \begin{subfigure}[b]{0.51\textwidth}
        \centering
        \includegraphics[width=1\linewidth]{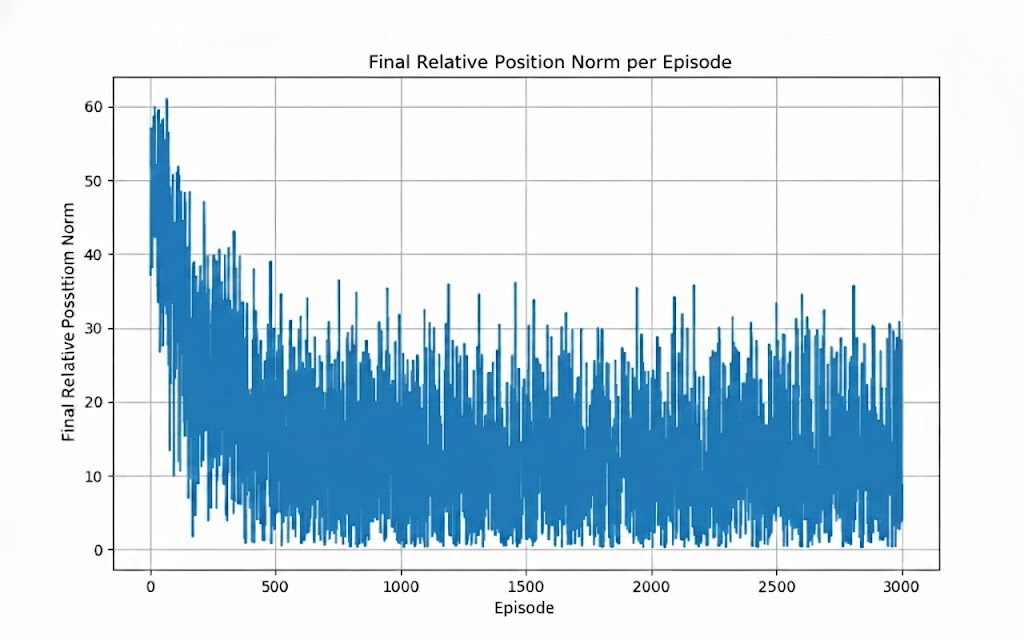}
        \caption{Norm of Final Relative Positions During Training}
        \label{fig:finalnorms_manual}
    \end{subfigure}
    \hfill
    \begin{subfigure}[b]{0.49\textwidth}
        \centering
        \includegraphics[width=1\linewidth]{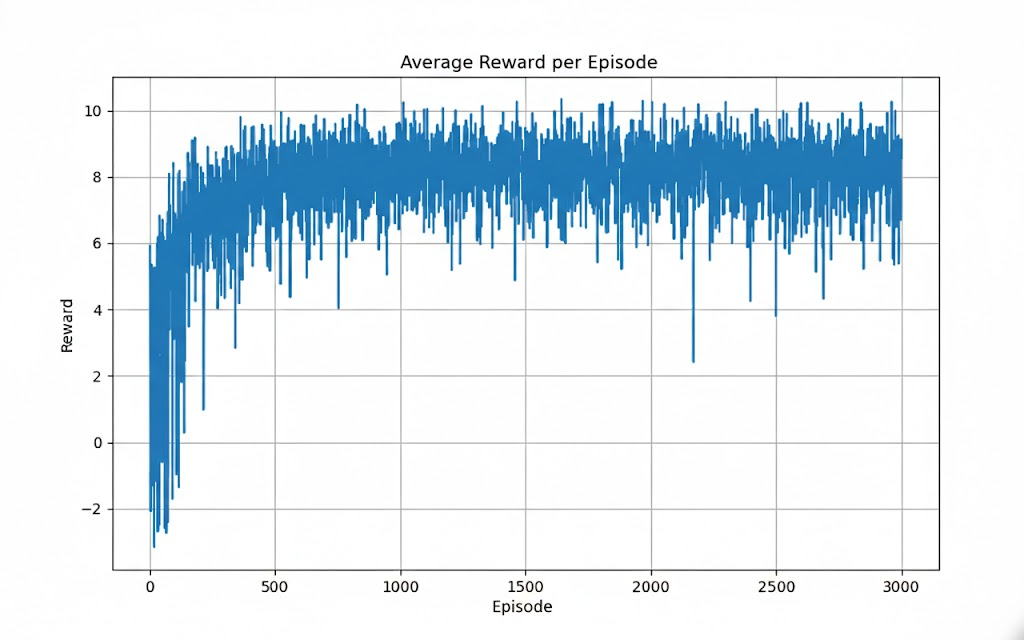}
        \caption{Reward Function During Training}
        \label{fig:rewards_manual}
    \end{subfigure}
    \caption{Docking Results from Manual Shaping and a 50m Initial Distance}
    \label{fig:Manual_Shaping_Results}
\end{figure}

The manual tuning process involved applying the process described in \ref{sec412} towards the tuning and configuration of a 3000-episode DDPG training. This training considered an initial separation of \(d_i=50m\) between the target and chaser spacecraft at the start of each episode and the representation of state, action, and reward provided in Section \ref{subsubsec:ddpg} except for the temporary relaxation of the velocity tolerance \(\epsilon_{vel}\) in the docking criteria and a position-only state vector ($\vec{s}_t=[\vec{x}_{t,target}-\vec{x}_{t,chaser}]$). \newline
\indent This process offered near-term potential as an option for an agent learning to reach the target. The complete DDPG training demonstrated improvement in reaching the target as shown in Fig.  \ref{fig:Manual_Shaping_Results}, with the magnitude of the final relative position decreasing during training and the agent learning to maximize rewards. The main limitation of the manual method is that even subtle modifications to the problem formulation require a high workload that must be repeated with each change. Examples of modifications that necessitate retuning include changes to the initial chaser-target separation, imposing terminal velocity constraints, or modifying the neural network architecture. The results in Fig. \ref{fig:Manual_Shaping_Results} demonstrate the limitations of manual tuning because the policy often reaches the general vicinity of the target without converging within the prescribed docking tolerance. Driving the relative position norms in Fig. \ref{fig:finalnorms_manual} to within the docking tolerance $\epsilon_{pos}$ demands significant manual shaping that often degrades learning stability; this motivates the use of Bayesian optimization which is an automated tuning method. \newline

\subsubsection{Automatic Tuning}
\begin{figure}[h!]
    \centering
    \begin{subfigure}[b]{0.49\textwidth}
        \centering
        \includegraphics[width=1\linewidth]{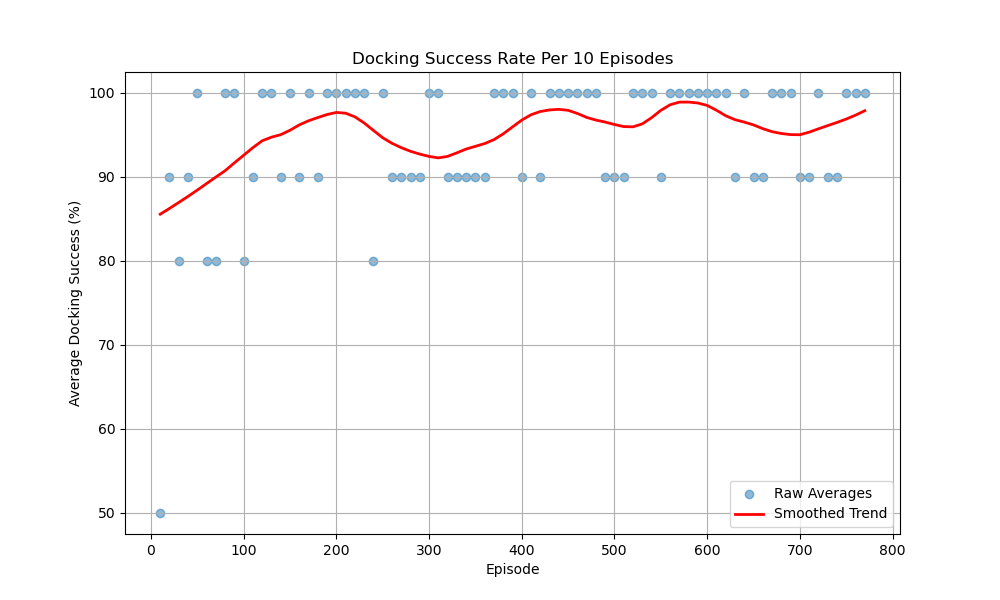}
        \caption{Iteration 7 (47 Successful Docks out of Final 50 episodes)}
        \label{fig:subfig1}
    \end{subfigure}
    \hfill
    \begin{subfigure}[b]{0.49\textwidth}
        \centering
        \includegraphics[width=1\linewidth]{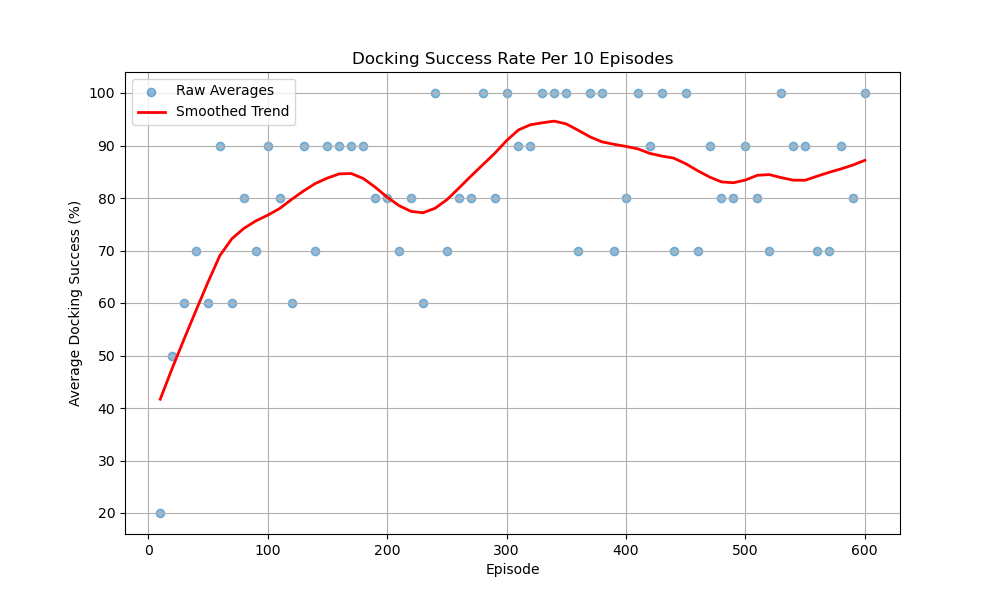}
        \caption{Iteration 8 (39 Successful Docks out of Final 50 episodes)}
        \label{fig:subfig2}
    \end{subfigure}
    
    \vspace{1em} 
    
    \begin{subfigure}[b]{0.49\textwidth}
        \centering
        \includegraphics[width=1\linewidth]{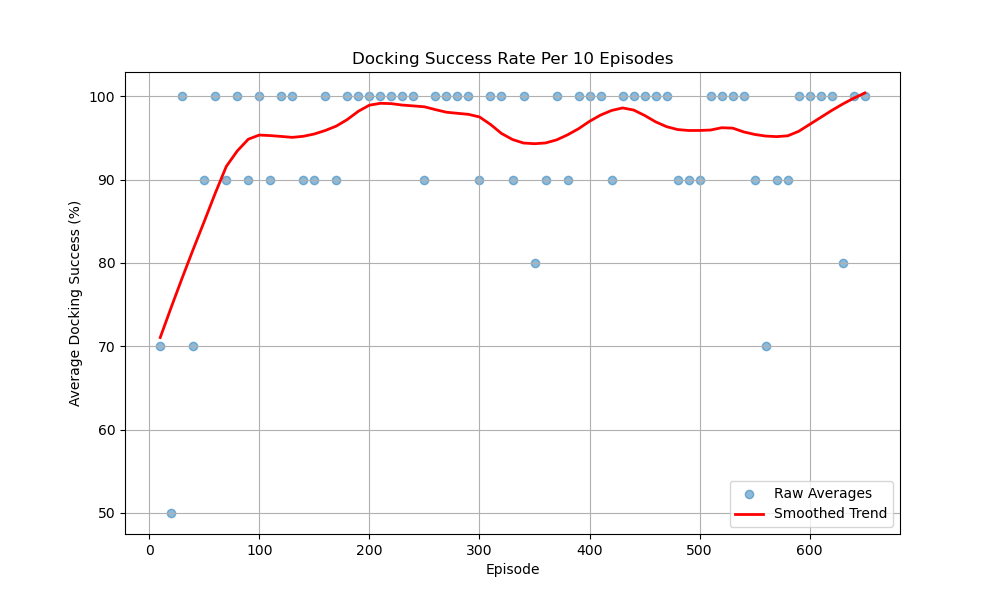}
        \caption{Iteration 13 (47 Successful Docks out of Final 50 episodes)}
        \label{fig:subfig3}
    \end{subfigure}
    \caption{Successful Iterations of Bayesian optimization}
    \label{fig:three-iter}
\end{figure}

\begin{figure}
    \centering
    \includegraphics[width=0.55\linewidth]{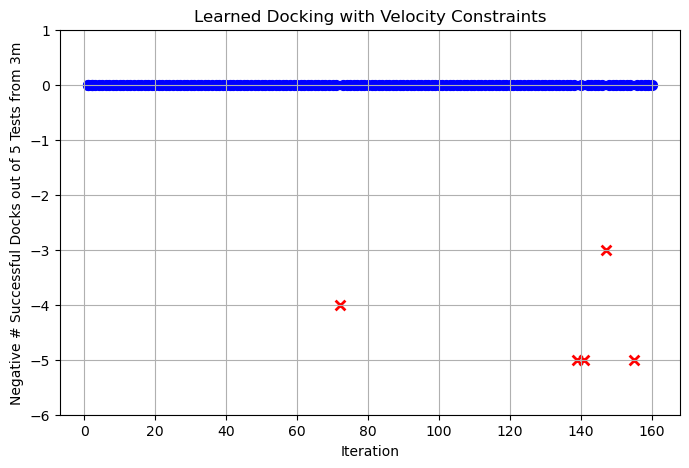}
    \caption{Bayesian optimization Trajectory with Velocity Constraints}
    \label{fig:bayesianopt_velconstraints}
\end{figure}

The automatic tuning process involved Bayesian optimization for hyperparameter tuning is described in \ref{alg_bayesianopt}. The objective function was formulated as the number of successful docks in the final 50 episodes of training, to be maximized by tuning all hyperparameters including learning rates, exploration rates, neural network architecture, and the total number of episodes. Each episode started with an initial separation between the chaser and target of \(d_i=10m\), and similar assumptions to those in Section \ref{manualtuningresults} were invoked. The representation of state, action, and reward are provided in Section \ref{subsubsec:ddpg}, except for the temporary relaxation of the velocity tolerance \(\epsilon_{vel}\) in the docking criteria criteria and a position-only state vector ($\vec{s}_t=[\vec{x}_{t,target}-\vec{x}_{t,chaser}]$).\newline
\indent When subject to this automatic tuning configuration, the actor/critic networks exhibited stable learning and demonstrated successful docking performance. The Bayesian optimization algorithm identified multiple reliable architectures capable of learning to dock at a success rate higher than 95\%. The top three RL trainings from a 20-iteration Bayesian optimization run are provided in Fig. \ref{fig:three-iter} where the ratio of successful docks is computed and plotted every ten episodes. \newline
 \newline
\indent The next automatic tuning configuration attempted to consider contact velocity as part of the docking criteria, which motivated the introduction of velocity in the state representation ($\vec{s}_t=[\vec{x}_{t,target}-\vec{x}_{t,chaser}, \vec{v}_{t,target}-\vec{v}_{t,chaser}]$) and the introduction of a velocity tolerance \(\epsilon_{vel}\) in the docking criteria criteria. The velocity term $r_{vel,t}$ in the reward function as formulated in Section \ref{subsubsec:ddpg} was included in this automatic tuning configuration.\newline
\indent The inclusion of this sparse velocity term was a pivotal improvement in docking performance. Many previous attempts involved using sparse penalties if the chaser exceeds the maximum relative contact velocity, which would be formulated in the configuration of Section \ref{subsubsec:ddpg} as:
\begin{align}
    r_{vel,t}=-c_3 \text{\ if\ } ||\textbf{\textit{s}}_t[1:3]||<\epsilon_{pos}\ \&\ ||\textbf{\textit{s}}_t[4:6]||<\epsilon_{vel}
\end{align}
where \(c_3\) is a tunable penalty constant. Agents in this formulation struggled to reach the target because the agent receives penalties while in the vicinity of the target, which inhibits the process of learning to approach the target. The convergence challenges that emerged from terminal velocity constraints were overcome by formulating sparse velocity rewards instead of sparse velocity penalties, where the learning algorithm incentivizes slower approaches rather than penalizing faster approaches. Learning a policy with velocity constraints is substantially more difficult, requiring closer initial separations, with the intent of employing curriculum learning to incrementally extend performance toward mission-ready initial distances.\newline
\indent A complete Bayesian optimization was employed to demonstrate the successful convergence to a policy that satisfies terminal velocity constraints. The optimization scheme evaluates each candidate policy post-training using five docking attempts from an initial distance \(d_i=3m\), seeking to maximize the number of successful docks with an objective function in the range \([-5, 0]\).  As shown in Fig. \ref{fig:bayesianopt_velconstraints}, multiple designs dock successfully all 5 times, exhibiting learned behavior that satisfies the maximum allowable relative contact velocity \(\epsilon_{vel}\).

\subsection{Navigation}

The navigation network is tested on the SPEED+ dataset without and with the augmentations described in Section \ref{sec:augments}. The comparison shown in Table \ref{tb:results_aug} demonstrates the effect of the inclusion of data augmentations, clearly showing the benefits of including data augmentation methods for bridging the domain gap. Table \ref{tb:speed+} shows the performance of SPN \citep{SPN}, KRN \citep{KRN}, and the MMEDR navigation network on the SPEED+ dataset. Most competitors in the SPEC 2021 competition utilize either adversarial or online training, meaning some amount of unlabeled testing data has been used for model tuning. Since these methods have not yet been implemented into the MMEDR-framework, a comparison to SPN and KRN is shown because these models do not contain online learning methods. This highlights the strengths of what work has been implemented in the navigation network. 

\begin{table}[h!bt]
    \centering
    \caption{Testing Results for Lightbox and Sunlamp Domains, With and Without Data Augmentation (Described in Section \ref{sec:augments})}
    \begin{tabular}{c@{\qquad}ccc@{\qquad}ccc}
      \toprule
      \multirow{2}{*}{\raisebox{-\heavyrulewidth}{Training}} & \multicolumn{3}{c}{Lightbox} & \multicolumn{3}{c}{Sunlamp} \\
      \cmidrule{2-7}
       Augments & $E_t$ (m) & $E_q$ ($^\circ$) & $E_p^*$ (-) & $E_t$ (m) & $E_q$ ($^\circ$) & $E_p^*$ (-) \\
      \midrule
      Without & 1.95 & 76.25 & 1.65 & 1.93 & 89.59 & 1.85 \\
      With & 0.36 & 45.65 & 0.86 & 0.52 & 66.74 & 1.28 \\
      \bottomrule
      \label{tb:results_aug}
    \end{tabular}
\end{table}

\begin{table}[h!bt]
    \centering
    \caption{Comparison of Synthetic-Only Training, Tested on Lightbox and Sunlamp Domains}
    \begin{tabular}{c@{\qquad}ccc@{\qquad}ccc}
      \toprule
      \multirow{2}{*}{\raisebox{-\heavyrulewidth}{Model}} & \multicolumn{3}{c}{Lightbox} & \multicolumn{3}{c}{Sunlamp} \\
      \cmidrule{2-7}
      & $E_t$ (m) & $E_q$ ($^\circ$) & $E_p^*$ (-) & $E_t$ (m) & $E_q$ ($^\circ$) & $E_p^*$ (-) \\
      \midrule
      SPN & 0.45 & 65.12 & 1.21 & 0.65 & 92.95 & 1.73 \\
      KRN & 2.25 & 44.53 & 1.12 & 14.64 & 80.95 & 3.73 \\
      Ours & 0.36 & 45.65 & 0.86 & 0.52 & 66.74 & 1.28 \\
      \bottomrule
      \label{tb:speed+}
    \end{tabular}
\end{table}

In relation to the SPEC 2021 competition \citep{SPECcomp}, the MMEDR navigation network performs right within the top ten in terms of $E_t$, despite the lack of heatmaps, adversarial training, or online training. However, the rotation prediction lags severely. Direct regression of orientation is difficult, especially for symmetric objects. Many networks in the competition use heatmap prediction for keypoint detection and PnP to gather orientation information, but it was avoided in this case to prioritize limiting model size and storage usage. However, because of these considerations, the average inference time on a single physical core (two threads) of an Intel Xeon w9-3475X Processor at 2.20 GHz (max 4.80 GHz) is 161.8 ms using Tensorflow \citep{tensorflow}, which results in an inference rate of 6.18 Hz. This network is extremely fast at only 11.2M parameters, showing room for improvement in terms of the base architecture. Additionally, this means that there is room to grow and add new tasks or include online training methods.

\subsubsection{Synthetic Data Generation}

The Aura spacecraft was chosen as the target and placed at a random distance from the surface of a scaled model of Earth; this distance was set to always be representative of an altitude between LEO and geostationary. The deep-space background was modeled as black, with no stars, Sun, or Moon present within the chaser spacecraft’s camera field of view. Blender was chosen for its ability to provide realistic renders with sophisticated lighting effects, along with the ability to script (and therefore automate) the image rendering process easily using Python. Using a script, the position and orientation of the camera relative to the target spacecraft can be randomized, as well as the target's altitude above the Earth and the time of day. This process was iterated to create a large dataset with minimal human intervention. Blender allows for object pose and other image metadata to be saved to a .json file, which is necessary for training any supervised machine learning model.  

The Earth model, rather than being a simple background image, is a full-scale representation of Earth, complete with 3D terrain, simulated atmosphere, city lighting at night, and movable clouds. The Earth's surface texture and cloud layers are rotated randomly for every image, allowing for greater variation. All of these elements combine to produce a more realistic Earth background that reflects the different effects the camera can see at any given point.

\indent The preliminary results of this process can be seen in Fig. 9. Although these images were generated prior to completing the full automation pipeline, they employed the desired Earth model, target model, and lighting affects, serving as a proof of concept for further data generation. Blender's in-house Cycles rendering engine was used to generate these images due to its physically-based ray-tracing framework. While this approach significantly increases computational cost and render time, it produces substantially more realistic lighting than Blender’s rasterized (OpenGL-based) EEVEE engine, which is the primary alternative within the software.

\begin{figure}[h!]
    \centering
    \begin{subfigure}[b]{0.49\textwidth}
        \centering
        \includegraphics[width=1\linewidth]{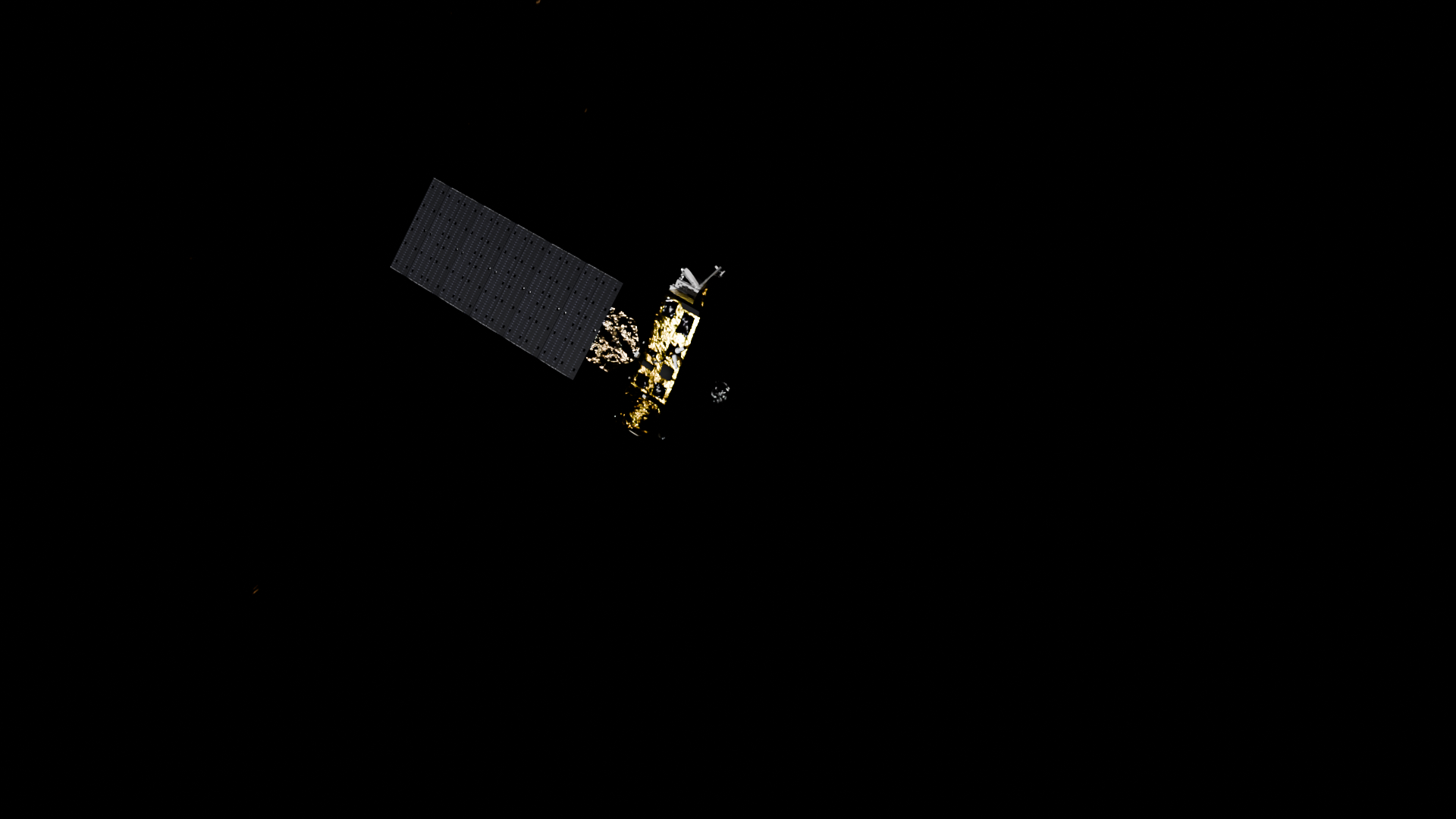}
        \label{fig:subfig1}
    \end{subfigure}
    \hfill
    \begin{subfigure}[b]{0.49\textwidth}
        \centering
        \includegraphics[width=1\linewidth]{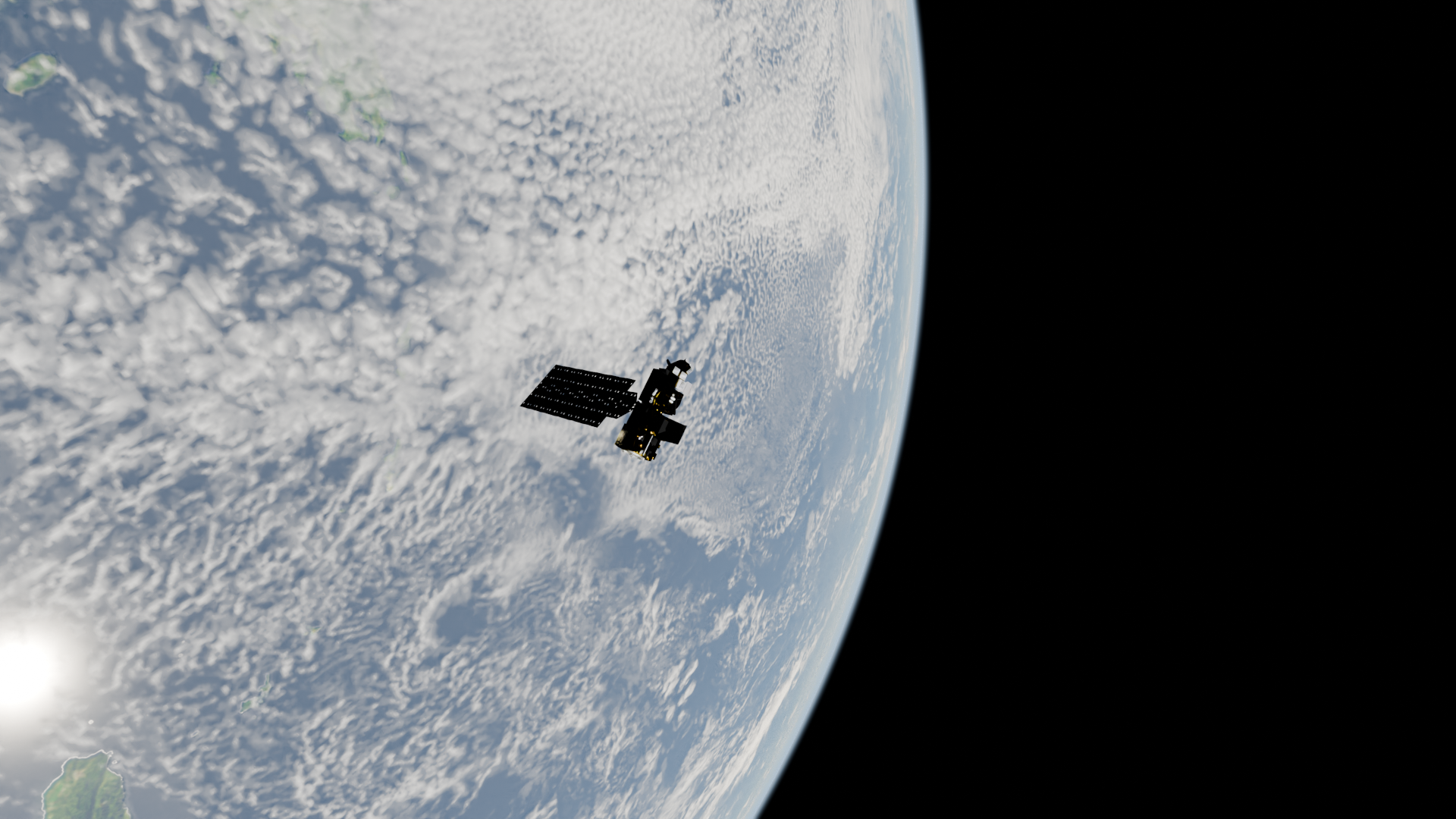}
        \label{fig:subfig2}
    \end{subfigure}
    \caption{Synthetic Images of the Aura Spacecraft Generated in Blender}
\end{figure}

\section{Conclusion and Future Work}\label{sec:conclusions}

The MMEDR-Autonomous framework shows promise for applications of ML-based GNC to the satellite rendezvous and docking problem. Several conclusions can be drawn from the analysis of multiple learned policies from DDPG; i) manual shaping can offer short-term results but Bayesian optimization is necessary for accomplishing sufficient performance in more complex problems, ii) the introduction of additional constraints substantially complicates the learning process and requires beginning with a simpler problem such as docking from a closer initial position when considering velocity constraints, and iii) sparse velocity rewards is a unique and valuable method to learn an acceptable contact velocity in contrast to existing literature that utilizes sparse velocity penalties. Future work on the guidance algorithms will involve the multi-agent problem formulation, the implementation of D4PG, and further unification of RL with optimal control methods. In terms of navigation, the creation of a fast and lightweight network often conflicts with the desire for accuracy in results. The MMEDR navigation network shows that networks using direct regression have the capability to compete with the traditional heatmap/keypoint or predetermined attitude class methods. Although data augmentation does aid in testing performance, other methods such as online or adversarial training are required to reach the accuracy levels necessary for deployment in flight. Future work includes investigating a variety of online training methods to improve testing performance without using excessive computational power. Additionally, a task specifically to aid in the docking process will be constructed, allowing the navigation network to identify "graspable" features on targets that may not have cooperative docking hardware. The development of the testing facility will serve as a reliable form of GNC ground testing. The current hardware for this laboratory mimics the imaging environment sufficiently for optical navigation testing and the robotic arms allow for guidance to move the simulated agent in real time. The continued construction of this facility includes connecting all of the individual components and calibrating the laboratory space to allow for the simulation of real-time rendezvous and generation of simulated images for navigation training. The GNC capabilities discussed in this work will be further improved and integrated together to create a single, fully autonomous agent that is verifiable in the on-ground DCV-Space laboratory.

\section*{Acknowledgments}
The authors acknowledge the financial support provided by Missouri University of Science and Technology through institutional seed funding for new faculty. This work was also supported in part by the Kummer Missouri S\&T Ignition Grant Initiative (IGI).

\appendix

\section{Derivation of Rotational \& Translational Relative Dynamics}\label{app:rotdynamics}
First, consider the following definitions:
\begin{itemize}
    \item $\mathcal{F}_{bs}$: Body frame of spacecraft $S$ (frame that is fixed to the body of chaser $S$)
    \item $\mathcal{F}_{bt}$: Body frame of spacecraft $T$ (frame that is fixed to the body of target $T$)
    \item $\omega_{bs}$ (vector notation dropped for simplicity): angular velocity of $S$ with respect to inertial frame $I$ 
    \item $\omega_{bt}$: angular velocity of $T$ with respect to inertial frame $I$
    \item $J_s, J_t$: mass moment of inertia matrices of $S$ and $T$ respectively, written in $3 \times 3$ matrix form with body-fixed axes
    \item $T_s$: external torque acting on $S$. It includes torque due to natural forces and torque due to its own control input
    \item $T_t$: external torque acting on $T$. It includes torque due to natural forces (and assumed to be unactuated)
    \item $q_r$: relative quaternion expressing the orientation of $\mathcal{F}_{bt}$ with respect to $\mathcal{F}_{bs}$, i.e., $q_r=q_t \otimes q_s^{-1}$ 
    \item $q_s$: relative quaternion expressing the orientation of $\mathcal{F}_{bs}$ with respect to inertial frame $I$
    \item $q_t$: relative quaternion expressing the orientation of $\mathcal{F}_{bt}$ with respect to inertial frame $I$
    \item $\omega_r$: angular velocity of $\mathcal{F}_{bt}$ with respect to $\mathcal{F}_{bs}$, i.e., $\omega_r = \omega_{bt} - \omega_{bs}$
    \item $A(q_r)$: Direction Cosine Matrix (DCM) or rotation matrix that converts a vector from $\mathcal{F}_{bs}$ to $\mathcal{F}_{bt}$
\end{itemize}
Recall the following fundamentals of quaternions:
\begin{align}
    q=\begin{bmatrix}
        q_1\\q_2\\q_3\\q_4
    \end{bmatrix}\equiv\begin{bmatrix}
        {q}_V\\q_4
    \end{bmatrix}
\end{align}
where \(q_1^2+q_2^2+q_3^2+q_4^2=1\)
\begin{align}
    q^{-1}=\begin{bmatrix}
        -q_V\\q_4
    \end{bmatrix}
\end{align}
\begin{align}
    q\otimes p=\begin{bmatrix}
        q_4\ P_V + P_4\ q_V - q_V\times P_V\\q_4 P_4 - q_V \cdot P_V
    \end{bmatrix}
\end{align}

The quaternion can be related to $A$ using the following identity \citep{spacecraftadc}:
\begin{align}
    A(q) = (q_4^2-q_V^Tq_V)I+2q_Vq_V^T-2q_4[q_V]_x
\end{align}
where the skew-symmetric matrix \([\cdot]_x\) is represented as:
\begin{align}
    [q_V]_x=\begin{bmatrix}
        q_1\\q_2\\q_3
    \end{bmatrix}_x=\begin{bmatrix}
        0 & -q_3 & q_2\\q_3 & 0& -q_1\\-q_2& q_1& 0\\
    \end{bmatrix}
\end{align}
Having defined the quaternion essentials, we now transition to the Euler's equation of motion (EOM) for rigid bodies:
\begin{align}
    J_s\ \dot{\omega}_{bs}^{(bs)}+\omega_{bs}^{(bs)}\times(J_s\ \omega_{bs}^{(bs)} )=T_s^{(bs)}\\
    J_t\ \dot{\omega}_{bt}^{(bt)}+\omega_{bt}^{(bt)}\times(J_t\ \omega_{bt}^{(bt)} )=T_t^{(bt)} 
\end{align}
where $(\cdot)^{(ij)}$ denotes the argument quantity being written using unit vectors of the $ij$ frame. Recalling that \(\omega_r=\omega_{bt}-\omega_{bs}\), when written using unit vectors of the \(F_{bt}\) frame: 
\begin{align}
    \omega_r^{(bt)}=\omega_{bt}^{(bt)}-\omega_{bs}^{(bt)} \implies \omega_r^{(bt)}=\omega_{bt}^{(bt)}-A\omega_{bs}^{(bs)}
\end{align}
where $A$ is the simplified notation for $A(q_r)$. Taking the time derivative with respect to an inertial frame:
\begin{align}
    \dot{\omega}_r^{(bt)}=\dot{\omega}_{bt}^{(bt)}-\frac{d}{dt}[A\omega_{bs}^{(bs)}]
\end{align}
\begin{align}
    \dot\omega_r^{(bt)} = \dot\omega_{bt}^{(bt)}+\omega_{bt}^{(bt)}\times A\omega_{bs}^{(bs)}-A\omega_{bs}^{(bs)}\times \omega_{bs}^{(bs)}-A\dot\omega_{bs}^{(bs)}
\end{align}
Now, using the identity $\dot{A}=-[\omega_{bt}^{(bt)}]_xA+A[\omega_{bs}^{(bs)}]_x$:
\begin{align}
    \dot\omega_r^{(bt)} = \dot\omega_{bt}^{(bt)}+[\omega_{bt}^{(bt)}]_x A\omega_{bs}^{(bs)}-A[\omega_{bs}^{(bs)}]_x\omega_{bs}^{(bs)}-A\dot\omega_{bs}^{(bs)}\end{align}
and since $[\omega_{bs}^{(bs)}]_x\omega_{bs}^{(bs)}=\omega_{bs}^{(bs)}\times\omega_{bs}^{(bs)}=0$, the following is true: $\dot\omega_r^{(bt)} = \dot\omega_{bt}^{(bt)}-A\dot{\omega}_{bs}^{bs}+[\omega_{bt}^{(bt)}]_xA\omega_{bs}^{bs}$. Next, upon rearrangement of the Euler’s EOM, one obtains:
\begin{align}
    \dot{\omega}_{bs}^{(bs)} = J_s^{-1}[T_s^{(bs)}-\omega_{bs}^{(bs)}\times(J_s\ \omega_{bs}^{(bs)})] \\
    \dot{\omega}_{bt}^{(bt)} = J_t^{-1}[T_t^{(bt)}-\omega_{bt}^{(bt)}\times (J_t\ \omega_{bt}^{(bt)})]
\end{align}
Therefore,
\begin{align}
    \dot\omega_r^{(bt)} = J_t^{-1}[T_t^{(bt)}-\omega_{bt}^{(bt)}\times (J_t\ \omega_{bt}^{(bt)})]-AJ_s^{-1}[T_s^{(bs)}-\omega_{bs}^{(bs)}\times(J_s\ \omega_{bs}^{(bs)})]+\omega_{bt}^{(bt)}\times A\omega_{bs}^{(bs)}\label{implies1}
\end{align}
Now, upon rearranging $\omega_r=\omega_{bt}-\omega_{bs}$ and writing all quantities using unit vectors of the $F_{bt}$ frame, one obtains:
\begin{align}
    \omega_{bt}^{(bt)} = \omega_r^{(bt)}+\omega_{bs}^{(bt)}=\omega_r^{(bt)}+A\omega_{bs}^{(bs)}\label{implies2}
\end{align}
Substituting \ref{implies2} in \ref{implies1} results in: 
\begin{align}
    \dot\omega_r^{(bt)}=J_t^{-1}[T_t^{(bt)}-(\omega_r^{(bt)}+A\omega_{bs}^{(bs)})\times J_t (\omega_r^{(bt)}+A\omega_{bs}^{(bs)})]-AJ_s^{-1}[T_s^{(bs)}-\omega_{bs}^{(bs)}\times(J_s\ \omega_{bs}^{(bs)})]+\omega_{bt}^{(bt)}\times A\omega_{bs}^{(bs)}
\end{align}
With the expression for $\dot{\omega}_r^{(bt)}$ established, the expression for $\dot{q}_r$ is given by \citep{appliedreachabilityanalysis}:
\begin{align}
    \dot q_r = \frac{1}{2}\Omega(\omega_r^{(bt)})q_r \\
\end{align}
where:
\begin{align}
    \Omega(\omega) = \begin{bmatrix}
        0 & \omega_z & -\omega_y & \omega_x \\
        -\omega_z & 0 & \omega_x & \omega_y \\
        \omega_y & -\omega_x & 0 & \omega_z \\
        -\omega_x & -\omega_y & -\omega_z & 0 \\
    \end{bmatrix}
\end{align}
Having derived the relative rotational dynamics in spacecraft body frames, the relative translational dynamics can be derived in the LVLH frame.\newline
\indent Let $L$ be the target LVLH frame as described in Section \ref{govdyn}, where \(\hat{r}\) is the vector from the center of Earth to the center of mass of the target, \(\hat{h}\) is the specific angular momentum vector for the target, and $\hat{\theta}$ completes the right-handed triad.
\begin{align}
    \vec\rho = \vec r_s - \vec r_t
\end{align}
where $\vec\rho$ is the relative position between the chaser and target, $\vec r_s$ is the inertial position vector of the spacecraft $S$ (chaser spacecraft), and $\vec r_t$ is the inertial position vector of the target spacecraft. The inertial acceleration of target spacecraft \(T\) is given as:
\begin{align}
    \ddot{\vec{r}}_t = \frac{^Id^2}{dt^2}(\vec r_t)=-\frac{\mu}{r_t^3}\vec r_t +\vec a_{p,t}
\end{align}
where $\mu$ is Earth's gravitational parameter and $\vec a_{p,t}$ is the resultant perturbing acceleration acting on spacecraft \(T\). Similarly, the inertial acceleration of spacecraft \(S\) is:
\begin{align}
    \ddot{\vec{r}}_s = \frac{^Id^2}{dt^2}(\vec r_s)=-\frac{\mu}{r_s^3}\vec r_s +\vec a_{p,s}+\vec a_{t,s}
\end{align}
where $\vec a_{p,s}$ is the resultant perturbing acceleration acting on spacecraft \(S\) and $\vec a_{t,s}$ is the thrust acceleration of spacecraft \(S\). Taking the double time derivative of $\vec \rho$ with respect to an observer fixed in the inertial frame:

\begin{align}
    \frac{^I d^2\rho}{dt^2}\equiv \ddot{\vec\rho}=\ddot{\vec{r}}_s-\ddot{\vec{r}}_t \\
    \ddot{\vec\rho}=-\frac{\mu}{r_s^3}\vec r_s +\vec a_{p,s}+\vec a_{t,s}+\frac{\mu}{r_t^3}\vec r_t -\vec a_{p,t}  \\
    \ddot{\vec\rho}=-\frac{\mu}{r_s^3}(\vec\rho+\vec r_t)+\vec a_{p,s}+\vec a_{t,s}+\frac{\mu}{r_t^3}\vec r_t-\vec a_{p,t}  \\
    \ddot{\vec\rho}=\frac{\mu}{r_t^3}\vec r_t\left(1-\frac{r_t^3}{r_s^3}\right)-\frac{\mu}{r_s^3}\vec\rho+\vec a_{p,s}+\vec a_{t,s}-\vec a_{p,t}  \\
    \ddot{\vec\rho}=\frac{\mu}{r_t^3}\vec r_t\left(1-\frac{r_t^3}{|\vec\rho+\vec r_t|^3}\right)-\frac{\mu}{r_s^3}\vec\rho+\vec a_{p,s}+\vec a_{t,s}-\vec a_{p,t}  \\
    \ddot{\vec\rho}=\frac{\mu}{r_t^3}\vec r_t\left(1-\frac{r_t^3}{(\rho^2+r_t^2+2\vec\rho \cdot \vec r_t)^{3/2}}\right)-\frac{\mu}{r_s^3}\vec\rho+\vec a_{p,s}+\vec a_{t,s}-\vec a_{p,t} \label{eqn3_rhoddot}
\end{align}
Let the parameter $q$ be defined as:
\begin{align}
    q = \frac{\rho^2+2\vec\rho \cdot \vec r_t}{r_t^2}
\end{align}
Now, the term inside parenthesis in Eq. \ref{eqn3_rhoddot} can be rewritten as:
\begin{align}
    1-\frac{r_t^3}{(\rho^2+r_t^2+2\vec\rho \cdot \vec r_t)^{3/2}}=1-\frac{r_t^3}{((1+q)r_t^2)^{3/2}} \\
    = 1-\frac{1}{(1+q)^{3/2}} \\
    = \frac{(1+q)^{3/2}-1}{(1+q)^{3/2}} \\
    =\frac{[(1+q)^{3/2}-1][(1+q)^{1/2}+1]}{(1+q)^{3/2}[(1+q)^{1/2}+1]} \\
    = \frac{(1+q)^2+(1+q)^{3/2}-(1+q)^{1/2}-1}{(1+q)^{3/2}[(1+q)^{1/2}+1]} \\
    = \frac{q^2+2q+(1+q)^{1/2}q}{(1+q)^{3/2}[(1+q)^{1/2}+1]} \\
    = \frac{q(q+2+(1+q)^{1/2})}{(1+q)^{3/2}[(1+q)^{1/2}+1]}
\end{align}
Therefore, \ref{eqn3_rhoddot} becomes:
\begin{align}
    \ddot{\vec{\rho}} =\frac{\mu}{r_t^3}\frac{q(q+2+(1+q)^{1/2})}{(1+q)^{3/2}[(1+q)^{1/2}+1]}\vec r_t-\frac{\mu}{r_s^3}\vec\rho+\vec a_{p,s}+\vec a_{t,s}-\vec a_{p,t}
\end{align}
Now, from transport theorem:
\begin{align}
    \frac{^Id}{dt}(\vec\rho)=\frac{^Ld}{dt}(\vec\rho)+\vec\omega_{L/I}\times \vec\rho
\end{align}
where $\frac{^Ld}{dt}(\vec\rho)$ is the rate of change of $\vec \rho$ as seen by an observer fixed in the LVLH frame and $\vec\omega_{L/I}$ is the angular velocity of LVLH frame with respect to the inertial frame. Now, taking the time derivative of the preceding equation with respect to an observer fixed in the inertial frame:
\begin{align}
    ^I\ddot{\vec{\rho}}=\frac{^Id}{dt}(^L\dot{\vec{\rho}} +\vec\omega_{L/I}\times \vec\rho) \\
    ^I\ddot{\vec{\rho}}=\frac{^Ld}{dt}(^L\dot{\vec{\rho}} +\vec\omega_{L/I}\times \vec\rho)+\vec\omega_{L/I}\times(^L\dot{\vec{\rho}} +\vec\omega_{L/I}\times \vec\rho) \\
    ^I\ddot{\vec{\rho}}= {}^{L}\ddot{\vec{\rho}} + {}^{L}{\dot{\vec\omega}}_{L/I}\times\vec\rho+\vec\omega_{L/I}
    \times {}^{L}{\dot{\vec\rho}}+\vec\omega_{L/I}
    \times {}^L {\dot{\vec\rho}}+\vec\omega_{L/I}\times(\vec\omega_{L/I}\times \vec\rho)
\end{align}
where ${}^L {\ddot{\vec\rho}}$ is the relative acceleration as seen by an observer fixed in the LVLH frame. Rearranging the preceding equation yields:
\begin{align}
    {}^L {\ddot{\vec\rho}}={}^I {\ddot{\vec\rho}}-2\vec\omega_{L/I}
    \times{}^L {\dot{\vec\rho}}-{}^L {\dot{\vec\omega}}_{L/I}\times\vec\rho-\vec\omega_{L/I}\times(\vec\omega_{L/I}\times \vec\rho)\\
    {}^L\ddot{\vec{\rho}} =\frac{\mu}{r_t^3}\frac{q(q+2+(1+q)^{1/2})}{(1+q)^{3/2}[(1+q)^{1/2}+1]}\vec r_t-\frac{\mu}{r_s^3}\vec\rho+\vec a_{p,s}+\vec a_{t,s}-\vec a_{p,t}-2\vec\omega_{L/I}\times{}^L\dot{\vec{\rho}}\\\notag-{}^L {\dot{\vec\omega}}_{L/I}\times\vec\rho-\vec\omega_{L/I}\times(\vec\omega_{L/I}\times \vec\rho)
\end{align}
Next, we can find $\vec\omega_{L/I}, \dot{\vec\omega}_{L/I}$ by propagating the absolute motion of the target using:
\begin{align}
    \ddot{\vec{r}}_t =-\frac{\mu}{r_t^3}\vec r_t +\vec a_{p,t}
\end{align}
While propagating the absolute motion of the target, the Cartesian state can be related to the corresponding Keplerian orbital elements through standard orbital mechanics equations. From the Keplerian elements, $\vec\omega_{L/I}$ can be obtained using:
\begin{align}
    \vec\omega_{L/I}=\begin{bmatrix}
        \dot\Omega sin(i)sin(\theta_t)+\dot{i}cos(\theta_t)\\
        \dot\Omega sin(i)cos(\theta_t)-\dot{i}sin(\theta_t)\\
        \dot\Omega cos(i)+\dot{\theta_t}
    \end{bmatrix}^T\begin{bmatrix}
        \hat{r}\\\hat{\theta}\\\hat{h}
    \end{bmatrix}
\end{align}
where $\Omega$ is the right ascension of the ascending node (RAAN), $i$ is the orbital inclination, and the argument of latitude $\theta_t$ is defined as the sum of the argument of perigee and the true anomaly, i.e., $\theta_t = \omega + \nu$. Furthermore, as presented in \cite{orbitalrendezvous_HPC}, the Gauss Variational Equations (GVE), in either their original formulation or modified forms, are expressed as:
\begin{align}
    \dot{i}=\frac{rcos(\theta_t)}{h}a_h\\
    \dot\Omega=\frac{rsin(\theta_t)}{hsin(i)}a_h\\
    \dot{\theta_t}=\sqrt{\frac{\mu}{p^3}}(1+ecos(\nu))^2-\frac{rsin(\theta_t)cos(i)}{hsin(i)}a_h \label{eqn:thetadott_gve}
\end{align}
In the classical formulation of the GVE, $(a_r, a_\theta, a_h)$ denote the components of the perturbing acceleration resolved along the $\hat{r}$, $\hat{\theta}$, and $\hat{h}$ directions, respectively.
As outlined in \cite{orbitalrendezvous_HPC}, ${}^L \dot{\vec{\omega}}_{L/I} = {}^I \dot{\vec{\omega}}_{L/I}$ is determined numerically by propagating the target’s absolute orbital dynamics over the time span of interest. From Cartesian coordinates, $\vec{\omega}_{L/I}$ can be calculated for a large number of time steps over the interval of interest, then a cubic spline (or any other higher order curve) can be fitted through each of $\vec{\omega}_{L/I}|_{x}, \vec{\omega}_{L/I}|_{y}, \vec{\omega}_{L/I}|_{z}$, and the fitted curve will effectively be a closed-form formula which can be differentiated with respect to time to find $\dot{\vec{\omega}}_{L/I}$\newline
\indent To summarize, the rotational kinematics and dynamics are described by the following 14-dimensional state:
\begin{align*}
\text{Rotational State} \;\rightarrow\; \{ q_s, \; \omega_{bs}^{(bs)}, \; q_r, \; \omega_r^{(bt)} \}
\end{align*}
where $q_s$ is the quaternion representing the orientation of $F_{bs}$ with respect to the inertial frame, $\omega_{bs}^{(bs)}$ is the angular velocity of $S$ with respect to inertial frame $I$ and expressed in the $S$ body frame unit vectors, $q_r$ is the relative quaternion used to represent orientation of $F_{bt}$ with respect to $F_{bs}$, and $\omega_{r}^{(bt)}$ is the angular velocity of $F_{bt}$ with respect to $F_{bs}$ and expressed using unit vectors of the $F_{bt}$ frame. The rotational state derivatives are given by:
\begin{align}
    \dot{q}_s=\frac{1}{2}\Omega(\omega_{bs}^{(bs)})q_s\\
    \dot{\omega}_{bs}^{(bs)}=J_s^{-1}[T_s^{(bs)}-\omega_{bs}^{(bs)}\times(J_s \omega_{bs}^{(bs)})]\\
    \dot{q}_r=\frac{1}{2}\Omega(\omega_{r}^{(bt)})q_r\\
    \dot\omega_r^{(bt)}=J_t^{-1}[T_t^{(bt)}-(\omega_r^{(bt)}+A\omega_{bs}^{(bs)})\times J_t (\omega_r^{(bt)}+A\omega_{bs}^{(bs)})]\\\notag-AJ_s^{-1}[T_s^{(bs)}-\omega_{bs}^{(bs)}\times(J_s\ \omega_{bs}^{(bs)})]+(\omega_{r}^{(bt)}+A\omega_{bs}^{(bs)})\times A\omega_{bs}^{(bs)}
\end{align}
The derived translational kinematics and dynamics are described by the following 12-dimensional state:
\begin{align*}
\text{Translational State} \;\rightarrow\; \{ \vec{\rho},\; {}^L\dot{\vec{\rho}},\; \vec{r}_t,\; \dot{\vec{r}}_t \}
\end{align*}
where $\vec\rho$ is the relative position of $S$ with respect to $T$ and expressed in target LVLH unit vectors, ${}^L\dot{\vec{\rho}}$ is the rate of change of $\vec\rho$ as seen by an observer fixed in the LVLH frame and expressed using LVLH unit vectors, $\vec{r}_t$ is the inertial position of the target spacecraft, and $\dot{\vec{r}}_t$ is the inertial velocity of the target spacecraft. The corresponding time derivatives are:
\begin{align}
    \frac{{}^L d\vec\rho}{dt}={}^L\dot{\vec{\rho}}\\
    {}^L\ddot{\vec{\rho}} =\frac{\mu}{r_t^3}\frac{q(q+2+(1+q)^{1/2})}{(1+q)^{3/2}[(1+q)^{1/2}+1]}\vec r_t-\frac{\mu}{r_s^3}\vec\rho+\vec a_{p,s}+\vec a_{t,s}-\vec a_{p,t}-2\vec\omega_{L/I}\times{}^L\dot{\vec{\rho}}\\\notag-{}^L {\dot{\vec\omega}}_{L/I}\times\vec\rho-\vec\omega_{L/I}\times(\vec\omega_{L/I}\times \vec\rho)\\
     \frac{{}^I d\vec{r}_t}{dt}=\dot{\vec{r}}_t\\
    {}^I\ddot{\vec{r}}_t =-\frac{\mu}{r_t^3}\vec r_t +\vec a_{p,t}
\end{align}

\bibliographystyle{jasr-model5-names}
\biboptions{authoryear}
\bibliography{refs}

\end{document}